\documentclass[lettersize,journal]{IEEEtran}
\usepackage{amsmath,amsfonts}
\usepackage{algorithmic}
\usepackage{algorithm}
\usepackage{booktabs} 
\usepackage{array}
\usepackage[caption=false,font=normalsize,labelfont=rm,textfont=rm]{subfig}
\usepackage{textcomp}
\usepackage{stfloats}
\usepackage{float}
\usepackage[ colorlinks = true,
linkcolor = blue,
urlcolor  = blue,
citecolor = red,
anchorcolor = green,]{hyperref}
\usepackage{pifont}
\newcommand{\blue}[1]{{#1}}
\usepackage{bbding}
\usepackage{verbatim}
\usepackage{utfsym}
\usepackage{graphicx}
\usepackage{epstopdf}
\usepackage{cite}
\usepackage{adjustbox}
\usepackage{float}
\usepackage{multirow}
\usepackage{makecell}
\usepackage{pifont}
\usepackage{amssymb}

\usepackage[table,xcdraw]{xcolor}
\usepackage[table,xcdraw]{xcolor}
\begin{document}

\title{EarthGPT-X: A Spatial MLLM for Multi-level Multi-Source Remote Sensing Imagery Understanding with Visual Prompting}



\author{IEEE Publication Technology,~\IEEEmembership{Staff,~IEEE,}
\thanks{This paper was produced by the IEEE Publication Technology Group. They are in Piscataway, NJ.}
\thanks{ }}

\markboth{}%
{Shell \MakeLowercase{\textit{et al.}}: A Sample Article Using IEEEtran.cls for IEEE Journals}

\author{
    Wei Zhang, Miaoxin Cai,~\IEEEmembership{Graduate Student Member,~IEEE,} Yaqian Ning, Tong Zhang,~\IEEEmembership{Graduate Student Member,~IEEE,}  Yin Zhuang\textsuperscript{\dag},~\IEEEmembership{Member,~IEEE,}  Shijian Lu\textsuperscript{\dag},~\IEEEmembership{Member,~IEEE,} \\ He Chen,~\IEEEmembership{Member,~IEEE,} Jun Li,~\IEEEmembership{Fellow,~IEEE}, and Xuerui Mao\textsuperscript{\dag}

    \thanks{This work was supported by the National Natural Science Foundation of China under Grant 92152109, in part by the General Program of the National Natural Science Foundation of China under Grant 62371048, and in part by the National Natural Science Foundation of China under Grant T2225019, and in part by the MOE Tier-2 project of Singapore under Grant MOE-T2EP20123-0003. (\textit{{\dag} Co-corresponding authors: Shijian Lu, Yin Zhuang, and Xuerui Mao})}
    \thanks{Wei Zhang is with the School of Interdisciplinary Science, Beijing Institute of Technology, Beijing 100081, China, and with the College of Computing and Data Science, Nanyang Technological University, Singapore (e-mail: w.w.zhanger@gmail.com).} 
    \thanks{Xuerui Mao is with Beijing Institute of Technology (Zhuhai), Zhuhai, 519088, China. State Key Laboratory of Explosion Science and Safety Protection, Beijing, 100081, China. (e-mail: maoxuerui@sina.com).}
    \thanks{He Chen, Yin Zhuang, Miaoxin Cai, and Tong Zhang are with the National Key Laboratory of Science and Technology on Space-Born Intelligent Information Processing, Beijing Institute of Technology, Beijing 100081, China. (e-mails: chenhe@bit.edu.cn, yzhuang@bit.edu.cn, 3120220667@bit.edu.cn, bit\_zhangtong@163.com).
    }
    \thanks{Yaqian Ning is with the School of Optics and Photonics, Beijing Institute of Technology, Beijing 100081, China.}
    \thanks{Shijian Lu is with the College of Computing and Data Science, Nanyang Technological University, Singapore. (e-mail: Shijian.Lu@ntu.edu.sg).}
    \thanks{Jun Li is with the School of Computer Science and Hubei Key Laboratory of Intelligent Geo-Information Processing, China University of Geosciences, Wuhan, 430078, China (e-mail: lijuncug@cug.edu.cn).
    }
}
\maketitle

\begin{abstract}

Recent advances in natural-domain multi-modal large language models (MLLMs) have demonstrated effective spatial reasoning through visual and textual prompting. However, their direct transfer to remote sensing (RS) is hindered by heterogeneous sensing physics, diverse modalities, and unique spatial scales. Existing RS MLLMs are mainly limited to optical imagery and plain language interaction, preventing flexible and scalable real-world applications. In this article, EarthGPT-X is proposed, the first flexible spatial MLLM that unifies multi-source RS imagery comprehension and accomplishes both coarse-grained and fine-grained visual tasks under diverse visual prompts in a single framework.
Distinct from prior models, EarthGPT-X introduces: 1) a dual-prompt mechanism combining text instructions with various visual prompts (i.e., point, box, and free-form) to mimic the versatility of referring in human life; 2) a comprehensive multi-source multi-level prompting dataset, the model advances beyond holistic image understanding to support hierarchical spatial reasoning, including scene-level understanding and fine-grained object attributes and relational analysis; 3) a cross-domain one-stage fusion training strategy, enabling efficient and consistent alignment across modalities and tasks. Extensive experiments demonstrate that EarthGPT-X substantially outperforms prior natural and RS MLLMs, establishing the first framework capable of multi-source, multi-task, and multi-level interpretation using visual prompting in RS scenarios. The code and dataset are available at \textit{https://github.com/wivizhang/EarthGPT-X}.

\end{abstract}


\section{Introduction}
\IEEEPARstart{T}{ypical} spatial understanding tasks in multi-modal learning require aligning image regions with text semantics\cite{you2023ferret}. 
Beyond holistic analysis, spatial tasks involve accurately interpreting specific regions or objects based on user instructions\cite{zhan2023rsvg,yuan2023osprey}. Most recently, enabling multi-modal large language models (MLLMs) with comprehensive spatial understanding has become the focus of research in natural scenarios\cite{zhang2023gpt4roi,zhou2023regionblip,chen2023shikra}. However, these models struggle with the remote sensing (RS) imagery, which features multi-source, multi-viewpoint, and scale variations due to different imaging conditions\cite{zhan2023rsvg}. Although existing RS MLLMs can handle multiple RS visual tasks\cite{zhang2024earthgpt,kuckreja2024geochat}, they primarily focus on holistic image understanding, lacking spatial perception. They solely rely on text instructions, which can lead to ambiguity and limited context. It is indispensable to develop a spatial-aware MLLM to interpret RS imagery under flexible interactions.

Notably, in the human world, people can achieve an all-around understanding of visual information, allowing for both zoom-in fine-grained analyses and zoom-out coarse-grained overviews\cite{zellers2019recognition}. Inspired by this, visual prompting models advance the field by injecting visual prompts beyond text instructions to enable multi-grained visual reasoning and enhance spatial perception\cite{jia2022visual}. Representative works include the promptable segmentation model Segment Anything (SAM)~\cite{kirillov2023segment}, 
which enables class-agnostic delineation from visual or text prompts and shows strong generalization as a versatile backbone. 
The region-of-interest model GPT4RoI~\cite{zhang2023gpt4roi} integrates large language models (LLMs) for localized reasoning, while Ferret~\cite{you2023ferret} 
focuses on fine-grained grounding, which can accept diverse region inputs, such as points and bounding boxes. Additionally, Wu et al. \cite{wu2024visual} provide a review of visual prompting approaches, including prompt generation and prompt learning. However, designed for natural images, these models cannot be directly adapted to the heterogeneous RS domain due to fundamental differences in sensing physics, modalities, and spatial resolutions.

\begin{figure*}[!t]
	\centering
		\includegraphics[scale=0.16]{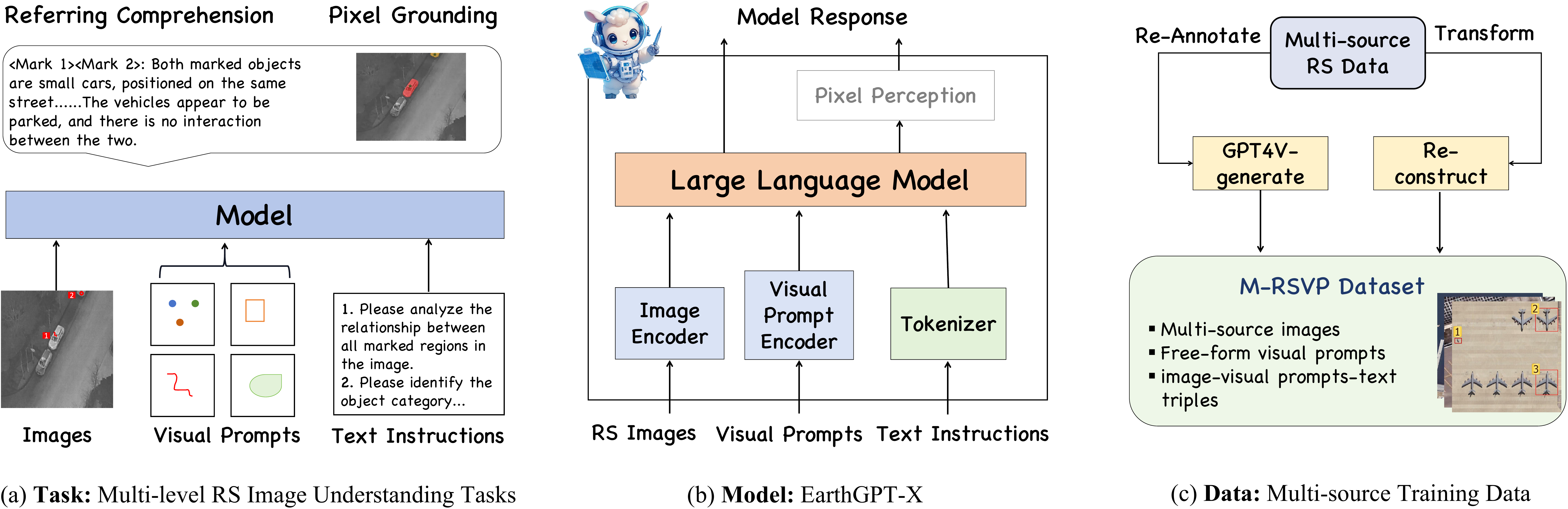}
	      \caption{Overview of the proposed framework for flexible and comprehensive spatial understanding in RS imagery: (a) \textbf{Tasks}: a wide range of multi-level visual reasoning tasks; (b) \textbf{Model}: EarthGPT-X integrates an image encoder, a visual prompt encoder, a text tokenizer, an LLM, and a pixel perception module; (c) \textbf{Data}: the training dataset M-RSVP is constructed from multi-source RS data (i.e., optical, synthetic aperture radar (SAR), infrared) and further enriched with GPT-4V–based annotations.}

	\label{FIG:overall}
\end{figure*}

Limited efforts in the RS domain have been made to enhance spatial understanding. A key example is RSPrompter~\cite{chen2024rsprompter}, which is based on SAM and develops an automated prompts generation approach to facilitate interactive segmentation. However, RSPrompter is only designed for instance segmentation. RSVG~\cite{zhan2023rsvg} defines the visual grounding tasks in the RS domain and can locate the referred objects using boxes under natural language expressions. Recent notable works, such as EarthGPT\cite{zhang2024earthgpt} and Geochat\cite{kuckreja2024geochat}, have realized RS holistic and regional imagery comprehension under the unified visual-language learning framework, facilitating the development of MLLMs in the RS domain. Nevertheless, these models solely support language interaction without visual prompts to realize referring expression comprehension. Notably, EarthMarker\cite{zhang2024earthmarker} is a visual prompting model in the RS domain, however, it lacks support for multi-source imagery and free-form visual prompts, hampering flexible human-AI interaction and comprehensive understanding of RS data. Thus, enabling MLLMs to interpret multi-source RS imagery at any granularity remains in its infancy.

To overcome these limitations, we propose EarthGPT-X, a spatial MLLM built upon visual prompt learning and tailored for multi-source RS imagery. It supports arbitrary visual prompt types (e.g., points, boxes, circles, and scribbles) and enables comprehensive interpretation from overall scene understanding to fine-grained regional analysis, thereby providing more flexible and scalable visual–language interaction than prior approaches. As depicted in Fig. \ref{FIG:overall}, the EarthGPT-X architecture consists of three interconnected components: multi-level RS image understanding \textit{tasks}, a visual prompting \textit{model}, and a construction pipeline of training \textit{dataset}. Specifically, EarthGPT-X not only allows for zoom-in fine-grained reasoning (e.g., free-form region captioning, referring object classification, and object inter-relationship analysis), but also supports zoom-out coarse-grained tasks (e.g., image-level captioning, scene classification). Secondly, the key of the model is to realize multi-modal content integration, including images, visual prompts, and text instructions. The multi-modal signals are then fed into the LLMs for a one-stage all-in-one training. Additionally, a pixel perception module is integrated to extend the model with object segmentation masks. Most importantly, the training dataset named M-RSVP is constructed, transforming from existing multi-sensor datasets, which are uniformly reorganized and relabeled as \textit{image-visual prompts-text} triples. In addition, a portion of the data re-annotated using GPT-4V\cite{achiam2023gpt} to generate longer and more detailed descriptions. A side-by-side comparison of EarthGPT-X with recent RS MLLMs is presented in Tab.~\ref{tab:tab1}, highlighting its advancement in modality coverage, dataset quality, prompting flexibility, and fine-grained multi-task reasoning.

To further validate EarthGPT-X's comprehensive interpretation ability on multi-source imagery, we compare it with recent state-of-the-art (SOTA) natural and RS MLLMs. The results indicate that EarthGPT-X achieves accurate multi-source reasoning and provides more detailed RS image understanding. Overall, EarthGPT-X not only attains competitive performance on visual–language benchmarks, but also demonstrates remarkable advantages in complex object-level reasoning tasks. 

Our contributions can be summarized as follows.
\begin{itemize}
\item{\textit{\textbf{EarthGPT-X: First Multi-source Spatial MLLM.}}
We propose EarthGPT-X, a spatial MLLM designed for multi-source (i.e., optical, SAR, and infrared) RS imagery understanding. It advances existing RS MLLMs by supporting free-form visual prompts (e.g., points, boxes, scribbles) to enable flexible multi-granularity interaction. EarthGPT-X unifies diverse spatial reasoning tasks into a visual prompting framework, providing a generalizable interpretation across heterogeneous RS data.}
\item{\textit{\textbf{M-RSVP: First Multi-source Visual Prompting Dataset.} }The first multi-source visual prompting dataset for RS, titled M-RSVP, consisting of over 650k \textit{image-visual prompt-text} triples, is constructed. The constructed M-RSVP bridges the gap of multi-source visual prompting data in the RS domain, enabling multi-granularity image interpretation across diverse scenes and sensors.

\item{\textit{\textbf{Multi-domain Multi-modal Multi-level Interaction}}. A hybrid signal mutual understanding mechanism is designed to strengthen the interplay between dense image features, sparse visual prompts, and text instructions, thereby improving the interpretation of referring regions within holistic images. Moreover, our cross-domain one-stage fusion training significantly streamlines the training procedure of MLLMs, while ensuring seamless integration of multi-modal knowledge.}

\item \textit{\textbf{Superior Performance.}} 
Extensive experiments benchmark EarthGPT-X against SOTA natural and RS MLLMs across diverse multi-level visual tasks and multi-source imagery, outperforming existing models. 
Furthermore, EarthGPT-X also excels in complex reasoning tasks, such as object inter-relationship analysis.
These advances collectively represent a significant milestone toward developing versatile and flexible MLLMs for RS scenarios.}
\end{itemize}

\begin{table*}[!t]
\caption{Comparisons of EarthGPT-X \textit{v.s.} Recent MLLMs.}
\centering
\scalebox{1.13}{
\begin{tabular}{c|c|cc|cc|cc|ccc|c}
\toprule
Model & Textual  & \multicolumn{2}{c|}{Modalities} & \multicolumn{2}{c|}{RS Dataset} & \multicolumn{2}{c|}{Tasks} & \multicolumn{3}{c|}{Visual Prompts} & Pix. \\
\cmidrule(r){3-4} \cmidrule(r){5-6} \cmidrule(r){7-8} \cmidrule(r){9-11}
 & Prompt & Opt. & Multi. & Pub. & Refi. & Regu. & Comp. & Point & Box & Free-form & Grd. \\
\midrule

Shikra~(\textit{arXiv 2023})\cite{chen2023shikra} &\Checkmark &\ding{55} &\ding{55} &\ding{55} &\ding{55} &\Checkmark&\Checkmark & \ding{55} & \Checkmark&\ding{55} &\ding{55} \\
GPT4RoI~(\textit{ECCV 2024})\cite{zhang2024gpt4roi} &\Checkmark &\ding{55} &\ding{55} &\ding{55} &\ding{55} &\Checkmark&\Checkmark & \ding{55} & \Checkmark&\ding{55} &\ding{55} \\
ViP-LLaVA~(\textit{CVPR 2024})\cite{cai2024vip} &\Checkmark &\ding{55} &\ding{55} &\ding{55} &\ding{55} &\Checkmark &\Checkmark & \Checkmark & \Checkmark&\Checkmark &\ding{55} \\
RS-LLaVA~(\textit{RS 2024})\cite{bazi2024rs} &\Checkmark &\Checkmark &\ding{55} &\Checkmark &\ding{55} &\Checkmark &\ding{55} & \ding{55} & \ding{55}&\ding{55} &\ding{55} \\
GeoChat~(\textit{CVPR 2024})\cite{kuckreja2024geochat} & \Checkmark&\Checkmark &\ding{55} &\Checkmark &\ding{55} &\Checkmark &\ding{55} & \ding{55} &\ding{55} &\ding{55} & \ding{55} \\
EarthGPT~(\textit{TGRS 2024})\cite{zhang2024earthgpt} &\Checkmark &\Checkmark &\Checkmark &\Checkmark &\ding{55} &\Checkmark &\ding{55} &\ding{55} & \ding{55} &\ding{55} & \ding{55} \\
Popeye~(\textit{JSTAR 2024})\cite{zhang2024popeye} &\Checkmark &\Checkmark &\Checkmark &\Checkmark &\ding{55} &\Checkmark &\ding{55} & \ding{55} & \ding{55} &\ding{55} & \Checkmark \\
LHRS-Bot~(\textit{ECCV 2024})\cite{muhtar2024lhrs} &\Checkmark & \Checkmark&\ding{55} &\Checkmark &\ding{55} &\Checkmark &\ding{55} &\ding{55} &\ding{55} & \ding{55} &\ding{55} \\
VHM~(\textit{AAAI 2025})\cite{pang2025vhm} &\Checkmark & \Checkmark&\ding{55} &\Checkmark &\Checkmark &\Checkmark &\ding{55} &\ding{55} &\ding{55} & \ding{55} &\ding{55} \\
EarthDial~(\textit{CVPR 2025})\cite{soni2025earthdial} &\Checkmark & \Checkmark&\Checkmark &\Checkmark &\ding{55} &\Checkmark &\ding{55} &\ding{55} &\ding{55} & \ding{55} &\ding{55} \\
EarthMarker~(\textit{TGRS 2025})\cite{zhang2024earthmarker} &\Checkmark & \Checkmark&\ding{55} &\Checkmark &\Checkmark &\Checkmark &\Checkmark & \Checkmark & \Checkmark &\ding{55} &\ding{55} \\
\textbf{EarthGPT-X (Ours)} &\Checkmark & \Checkmark&\Checkmark &\Checkmark &\Checkmark &\Checkmark &\Checkmark & \Checkmark &\Checkmark & \Checkmark& \Checkmark \\
\bottomrule
\end{tabular}
}
\label{tab:tab1}
\begin{flushleft}
\scriptsize \textbf{Note:} Comparisons in terms of modalities (\textit{\textbf{opt}ical or \textbf{multi}-source}), dataset construction (\textit{converted directly from \textbf{pub}lic data or \textbf{Refi}ned using powerful MLLMs such as GPT-4V or Gemini}), supported tasks (\textit{\textbf{Regu}lar visual tasks or \textbf{comp}lex reasoning such as inter-relationship analysis}), visual prompting capabilities, and extended pixel-level grounding.
\end{flushleft}
\end{table*}

\section{Related Work}

\subsection{MLLMs in Natural Scenarios}
The rapid development of LLMs has significantly advanced natural language processing (NLP), demonstrating remarkable capabilities in diverse contexts~\cite{brown2020language,openai2023chatgpt,zhang2023llama}. Then, an essential step toward achieving general intelligence is integrating multi-modal perception into LLMs, allowing them to handle multi-modal data beyond text. Several works~\cite{chen2022visualgpt,li2023blip,liu2023visual,alayrac2022flamingo} focus on incorporating multi-modal inputs to support task-specific objectives across various modalities. Models like VisualGPT~\cite{chen2022visualgpt}, Flamingo~\cite{alayrac2022flamingo}, and SPHINX-X\cite{pmlr-v235-liu24cc} have shown strong multi-modal reasoning abilities by aligning LLMs with visual data. This alignment is achieved through projection layers, zero-shot attention, and intermediate networks, as in MiniGPT-4~\cite{zhu2023minigpt}, 
LLAMA-Adapter V2~\cite{gao2023llama}, and mPLUG-Owl~\cite{ye2023mplug}. 
More recent advances include Qwen2.5-VL~\cite{qwen2.5-VL}, MiniCPM-V2.6~\cite{yao2024minicpm}, InternVL2.5~\cite{chen2024expanding}, InternLM-XComposer2.5~\cite{internlmxcomposer2_5_reward}, and DeepSeek-VL2~\cite{wu2024deepseekvl2mixtureofexpertsvisionlanguagemodels}. These enhanced versions advance prior frameworks by broadening input modalities, leveraging higher-quality large-scale datasets, and extending the scope of multi-modal reasoning tasks.

\subsection{Specialist and MLLMs in RS Scenarios}

RS imagery encompasses diverse modalities such as SAR, optical, and infrared, each capturing complementary physical characteristics of the observed scene. Recent advanced SAR detection adopts generative or predictive paradigms, e.g., MaDiNet~\cite{zhou2025madinet} with Gamma Diffusion–based bounding-box generation and SAR-JEPA~\cite{li2024predicting} with joint-embedding predictive recognition. In optical understanding, specialist models such as CSDS~\cite{wang2021csds} are designed for scene classification, while Qonly~\cite{marino2019ok} targets VQA tasks. For infrared detection, IRSAM~\cite{zhang2024irsam} enhances SAM to achieve more discriminative representations. Although effective for their respective tasks, these specialist models lack the ability to perform multiple visual tasks in a unified architecture, thus limiting scalability and hindering broader real-world deployment. 

Inspired by the success of MLLMs in natural image domains, RS research has also begun to adopt MLLMs with notable progress.
RSGPT~\cite{hu2023rsgpt}, built on Instruct-BLIP~\cite{instructblip}, demonstrates strong performance in captioning and VQA but struggles with classification and detection.
Cross-modal transfer learning in the RS MLLM field has recently gained traction, 
with representative works exemplifying different stages of progress \cite{li2023llava,ou2025geopix}. 
GeoChat~\cite{kuckreja2024geochat} represents the first versatile RS MLLM for multiple optical tasks, 
while SkyEyeGPT~\cite{zhan2025skyeyegpt} further extends broader task coverage, EarthGPT~\cite{zhang2024earthgpt} 
bridging MLLMs with optical, SAR, and infrared domains. 
LHRS-Bot~\cite{muhtar2024lhrs} additionally contributes by introducing LHRS-Bench for standardized evaluation.
These efforts are collectively paving the way for more unified and versatile frameworks. However, they remain limited to text-only prompts, lacking flexible interactive comprehension under visual instructions.

\subsection{Prompting Learning}

Prompt learning has become a key research direction in NLP~\cite{brown2020language}, with works such as AutoPrompt~\cite{shin2020autoprompt} and CoOp~\cite{zhou2022learning} automating prompt design for language and visual–language models. Subsequently, the idea evolved into visual prompting~\cite{wu2024visual}, marking a broader shift from language-based to multi-modal prompts. A milestone in this line is the SAM~\cite{kirillov2023segment}, which enables class-agnostic segmentation via point, box, or text prompts and demonstrates strong generalization across diverse domains.
GPT4RoI~\cite{zhang2023gpt4roi} and RegionBlip~\cite{zhou2023regionblip} leverage spatial boxes and combine language with region-of-interest input for regional recognition. Colorful Prompting Tuning (CPT)~\cite{yao2022cptcolorfulprompttuning} uses color-based prompts to enhance MLLMs' performance. Osprey~\cite{yuan2023osprey} goes further by incorporating fine-grained mask regions into language instructions for pixel-level understanding. ViP-LLava\cite{li2023llava} supports arbitrary
visual cues like scribbles and arrows to interpret images flexibly. Wu et al.~\cite{wu2024visual} provide a comprehensive survey of visual prompting approaches, 
covering visual prompting, prompt generation, and prompt learning. These models are primarily designed for the natural image domain and fail to generalize to the RS multi-source data. Notably, EarthMarker~\cite{zhang2024earthmarker} is a visual prompting framework tailored for the RS data, but it remains confined to optical imagery and does not support flexible prompt types. These gaps highlight the necessity of a specialized model that can support multi-source imagery and flexible interaction in RS scenarios, motivating the development of the proposed model.

\section{Methodology}
The EarthGPT-X framework is detailed in this section. We first elaborate on the multi-modal content integration mechanism (Section III-A). Then, the cross-domain one-stage fusion training strategy is presented (Section III-B). The extended pixel perception module is introduced in Section III-C.

\subsection{Multi-modal Content Integration} 

\begin{figure*}[!t]
	\centering
		\includegraphics[scale=0.135]{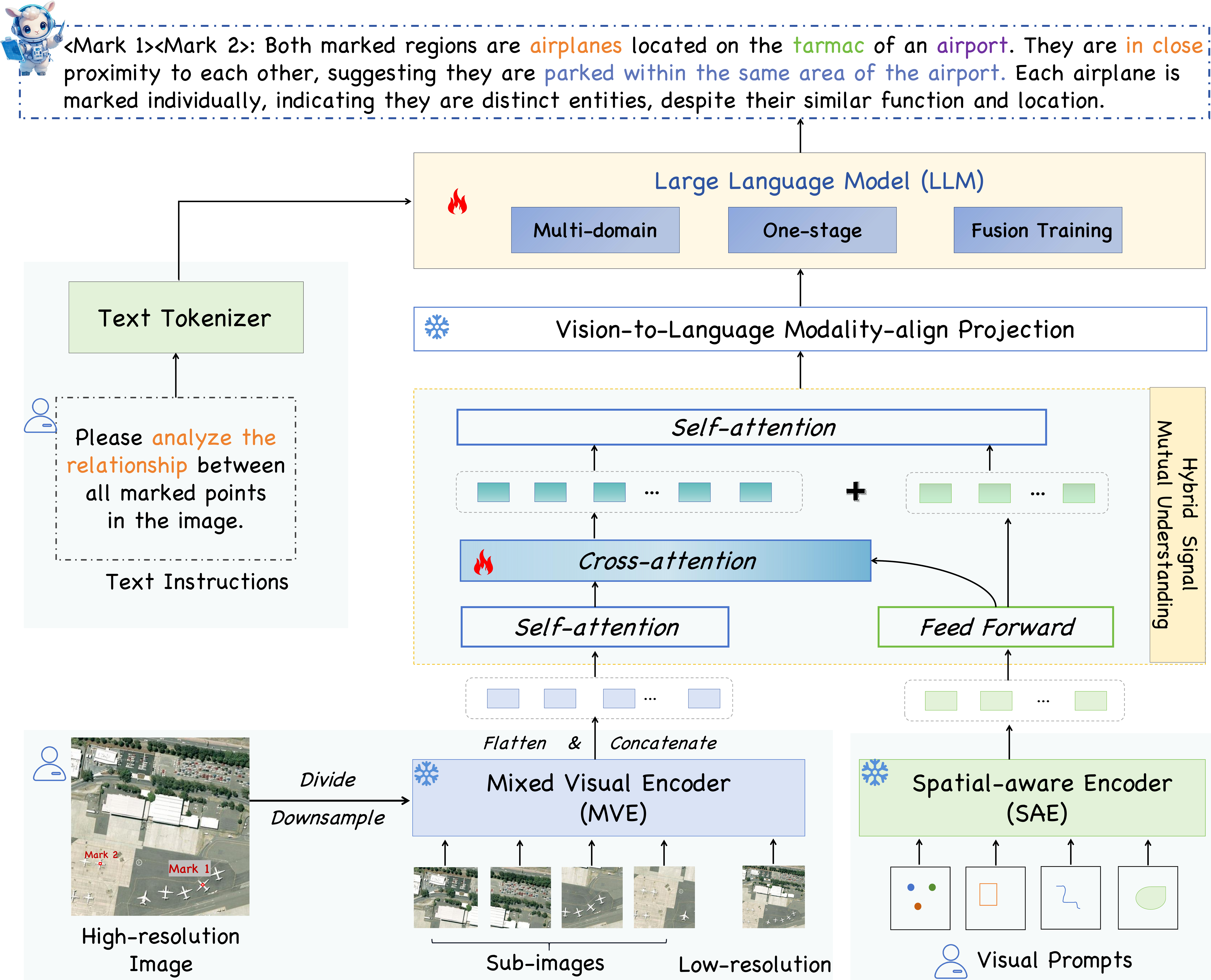}
	\caption{Overview of the proposed EarthGPT-X architecture. It integrates images, visual prompts, and text instructions through a hybrid signal interaction mechanism, and employs a unified one-stage fusion training strategy to enable multi-source knowledge integration and multi-level multi-task reasoning.}
	\label{FIG:model}
\end{figure*}

The multi-modal input stream of the EarthGPT-X is characterized by its tripartite structure. The multi-modal content encoding and integration are detailed as follows.

\subsubsection{Visual Global and Local Perception} As shown at the bottom left of Fig. \ref{FIG:model}, to enhance visual comprehension, the high-resolution input images are divided into sub-images for local visual perception. Concurrently, the images are downsampled into lower resolutions and combined with the sub-images for subsequent mixed encoding. The Mixed Visual Encoders (MVE) are designed to encode input images and their variants $I^i$ into visual representations $V_\mathrm{img}$. To leverage the advantages of different network architectures (CNN and Vision Transformer (ViT) \cite{dosovitskiy2021imageworth16x16words}) and integrate the local information and contextual dependencies, the semantically complementary backbones CLIP-ConvNeXt \cite{radford2021learning} and ViT self-supervised by DINOv2\cite{oquab2023dinov2} are employed as MVE. This visual encoding process can be formulated as 
\begin{equation}
V_\mathrm{img}=\mathrm{Concat} ~(\mathrm{MVE}~(I^i)),~i = 1,2,...,N.
\end{equation}

\subsubsection{Free-form Spatial-aware} Free-form visual prompts are expected to be flexibly used to guide the model in generating appropriate responses. The shapes of visual prompts include points, boxes, circles, scribbles, etc. To meet the need for flexible interaction at any granularity, a noise augmentation method is developed. Specifically, Gaussian distribution based on the dimensions of the box prompts is sampled to simulate the free-formed shapes. The introduced noise encompasses only a portion of the ground truth or expands it. Subsequently, considering that the visual prompts exhibit varying degrees of sparsity across different shapes, as illustrated in the bottom right of Fig. \ref{FIG:model}, a randomly initialized ViT is employed as a spatial-aware encoder (SAE) specifically for feature extraction, without any pre-trained weights suitable for image features. The various types of visual prompts $V_\mathrm{p}$ are encoded using the designed SAE. The implementation equation can be expressed as follows
\begin{equation}
E_\mathrm{prompt}=\mathrm{SAE}~(V_\mathrm{p}), 
\end{equation}
where $E_\mathrm{prompt}$ denotes the encoded visual prompts features with free-form shapes.

\subsubsection{Text Instructions Tokenizer} As displayed in left of Fig. \ref{FIG:model}, to facilitate linguistic comprehension, the text instructions $T_\mathrm{orig}$ are encoded by the tokenizer\cite{kudo2018sentencepiece} module, and projected into language embeddings $L_\mathrm{instruct}$, preparing for the subsequent LLM to capture the nuanced interdependence. The detailed process after disassembly is first to use the subword tokenization\cite{kudo2018sentencepiece} method to divide text instructions into subwords, enabling understanding of the structure of the linguistic signals. Subsequently, the Word2Vec\cite{mikolov2013efficient} is adopted to map each subword to a one-dimensional vector embedding. The tokenizer process can be simply described as
\begin{equation}
L_\mathrm{instruct}=\mathrm{Tokenizer}(~T_\mathrm{orig}).
\end{equation}

\subsubsection{Hybrid Signals Mutual Understanding} To bridge the gap and enhance the interplay between dense visual representations and sparse visual prompts features, a hybrid signal mutual understanding mechanism is proposed. Specifically, as displayed in the middle of Fig. 
\ref{FIG:model}, the image representations $V_\mathrm{img}$ and visual prompts features $E_\mathrm{prompt}$ are processed by a self-attention layer and feed-forward layer to obtain $V_\mathrm{sa}$ and $E_\mathrm{ff}$, respectively. The subsequent cross-attention and second self-attention blocks are used to compute the correlation $H_\mathrm{ca}$ between hybrid referring information and multi-scale holistic signals. In essence, this designed mechanism enhances detailed and contextual visual perception, leading to a deeper comprehension of the referring areas in holistic images. The whole procedure can be mathematically represented as
\begin{align}
V_\mathrm{sa} &=\mathrm{Self\text{-}attention}~ (V_\mathrm{img} ),\\
E_\mathrm{ff} &=\mathrm{Feed\text{-}forward}~ (E_\mathrm{prompt} ),\\
H_\mathrm{ca} &=\mathrm{Cross\text{-}attention}~ (V_\mathrm{sa}, ~E_\mathrm{ff}),\\
[V\text{-}P]_\mathrm{hybrid} &=\mathrm{Self\text{-}attention}~ (H_\mathrm{ca}+E_\mathrm{ff}),
\end{align}
where $[V\text{-}P]_\mathrm{hybrid}$ is the output of this designed module, containing rich visual information and spatial-aware knowledge from the free-form visual prompts. Then, the hybrid visual features are projected to the same dimension as text embeddings through the vision-to-language modality-align projection layer $\mathcal{F}_\mathrm{vl\text{-}p}$. This process can be formulated as
\begin{equation}
[V\text{-}P]_\mathrm{proj}=\mathcal{F}_\mathrm{vl\text{-}p}~([V\text{-}P]_\mathrm{hybrid}).
\end{equation}
where $[V\text{-}P]_\mathrm{proj}$ involve both multi-scale visual signals and free-form spatial-aware information. Notably, the cross-attention layer is updatable during training. 

\subsection{Cross-domain One-Stage Fusion Training} 
The conventional training pipeline of MLLMs typically follows a two-stage approach~\cite{zhang2024earthgpt,zhang2024earthmarker}, which requires extensive manual intervention. In particular, researchers must carefully configure trainable parameters and dataset combinations across different stages, making the process complex and labor-intensive. Different from the prior models, we adopt a cross-domain one-stage fusion training strategy~\cite{pmlr-v235-liu24cc}, which streamlines the process by uniformly treating all collected datasets while preserving strong performance. More importantly, this unified paradigm facilitates mutual understanding across heterogeneous domains, including natural, optical, SAR, and infrared imagery.

In this stage, the projected hybrid tokens are concatenated with the text instruction embeddings $L_\mathrm{instruct}$ to develop the multi-modal LLM input $\mathcal X$. Subsequently, the integrated multi-modal input is injected into LLM for alignment and deeply interactive understanding (see Fig. \ref{FIG:model} LLM part). The process can be expressed as
\begin{equation}
Y= \mathrm{LLM}~ ( ~[V\text{-}P]_\mathrm{proj},~L_\mathrm{instruct}).
\end{equation}
Unlike other models that introduce additional learnable parameters, to simplify the model structure, the Transformer's attention layers are unfrozen during the training. Specifically, the self-attention head consists of key $K$, query $Q$, and value $V$, which are implemented by linear layers. The implementations can be expressed as follows
\begin{equation}
Q(\mathcal X) = W_\mathrm{q} ~\mathcal X +{b_\mathrm{q}},
\end{equation}
\begin{equation}
K(\mathcal X) = W_\mathrm{k} ~\mathcal X +{b_\mathrm{k}},
\end{equation}
\begin{equation}
V(\mathcal X) = W_\mathrm{v} ~\mathcal X +{b_\mathrm{v}},
\end{equation}
where $\mathcal X$ denotes multi-modal input. The parameters $W_\mathrm{q}$, $W_\mathrm{k}$, $W_\mathrm{v}$, $b_\mathrm{q}$, $b_\mathrm{k}$, and $b_\mathrm{v}$ are updatable during the training. 

To summarize, EarthGPT-X leverages multi-domain knowledge to generalize versatile visual reasoning into multi-source RS scenarios. 
By tuning on abundant unified dual-prompting data, it achieves flexible referring comprehension with free-form visual prompts and text instructions.

\subsection{Integrated with Pixel Perception Module} 
To further enhance spatial perception, we integrate a pixel perception module into EarthGPT-X, enabling pixel-level grounding under flexible visual prompts. 
In particular, the special token $pix$ is adopted to represent the pixel grounding masks as embeddings\cite{he2024multi}. The embeddings $L_\mathrm{pix}$ are generated from the LLM output corresponding to the $pix$ token. Then, the language-to-pixel projection layer $\mathcal{R}_\mathrm{lp\text{-}p}$ transforms the embeddings $L_\mathrm{pix}$ into the pixel decoder's feature space. In addition, the grounding encoder is denoted as $\mathcal G$, the pixel decoder $\mathcal{P} $. The binary segmentation mask $M$ generated process can be represented as follows
\begin{equation}
M=\mathcal{P} \Big(\mathcal{R}_\mathrm{lp\text{-}p}(L_\mathrm{pix}),\mathcal{G} (I)\Big).
\end{equation}

In summary, EarthGPT-X, built on a tripartite architecture and an all-in-one training strategy, enables effective cross-modal and multi-source knowledge comprehension. 
It further supports multi-grained RS visual tasks with flexible integration of referring and grounding in a unified framework.

\begin{table}[!b]
\centering
\renewcommand{\arraystretch}{1.5}
\caption{Statistics of Training Data.}
\label{tab:Dataset}
\setlength{\tabcolsep}{0.7pt}
\scalebox{0.76}{
\fontsize{12pt}{12pt}\selectfont
\begin{tabular}{ccc@{\hskip 7.5pt}c}
\hline
Tasks & Raw Data & \#Samples & Domain \\ \hline
\multirow{2}{*}{Image Captioning}
& RSICD~\cite{lu2017exploring} & 24.3k & \multirow{6}{*}{\centering Optical} \\
& UCM-Captions~\cite{qu2016deep} & 10k & \\ \cline{1-3}
\multirow{2}{*}{Scene Classification}
& RSSCN7~\cite{zou2015deep} & 0.56k & \\
& EuroSAT~\cite{helber2019eurosat} & 5.4k & \\ \cline{1-3}
ROC (point) & SIOR~\cite{wang2023samrs} & 200k & \\ \cline{1-3}
Region Captioning & DIOR-RSVG~\cite{zhan2023rsvg} & 31.5k & \\ \hline
\multirow{6}{*}{ROC (box)}
& AIR-SARShip~\cite{xian2019air} & 1.43k & \multirow{6}{*}{\centering SAR} \\
& SSDD~\cite{zhang2021sar} & 1.86k & \\
& SARDet~\cite{li2024sardet} & 106k & \\
& SAR Aircraft~\cite{zhirui2023sar} & 1.9k & \\
& SAR-Ship~\cite{rs11070765} & 33.7k & \\
& MSAR~\cite{chenlarge} & 32.5k & \\ \hline
\multirow{4}{*}{ROC (box)}
& Sea-shipping\cite{Sea-shipping} & 8.4k & \multirow{4}{*}{\centering Infrared} \\
& Infrared-security~\cite{Infrared-security} & 9.6k & \\
& Aerial-mancar~\cite{InfiRay-Aerial-mancar} & 31.6k & \\
& Double-light-vehicle~\cite{Double-light-vehicle} & 5.1k & \\ \hline
\multirow{6}{*}{\makecell{Captioning,\\Relationship Analysis\\\\\textit{(GPT-assisted)}}}
& AIR-SARShip~\cite{xian2019air} & 0.5k & \multirow{4}{*}{\centering SAR} \\
& SARDet~\cite{li2024sardet} & 106k & \\
& SSDD~\cite{zhang2021sar} & 1.0k & \\
& SAR Aircraft~\cite{zhirui2023sar} & 1.9k & \\ \cline{2-4}
& Infrared-security~\cite{Infrared-security} & 9.6k & \multirow{2}{*}{\centering Infrared} \\
& Aerial-mancar~\cite{InfiRay-Aerial-mancar} & 31.6k & \\ \hline
\end{tabular}}
\begin{flushleft}
\scriptsize
\textit{Notes: ‘ROC’ denotes Referring Object Classification. The ‘box’ and ‘point’ refer to box-level and point-level visual prompts, respectively.}
\end{flushleft}
\end{table}

\begin{figure*}[!t]
	\centering
		\includegraphics[scale=0.19]{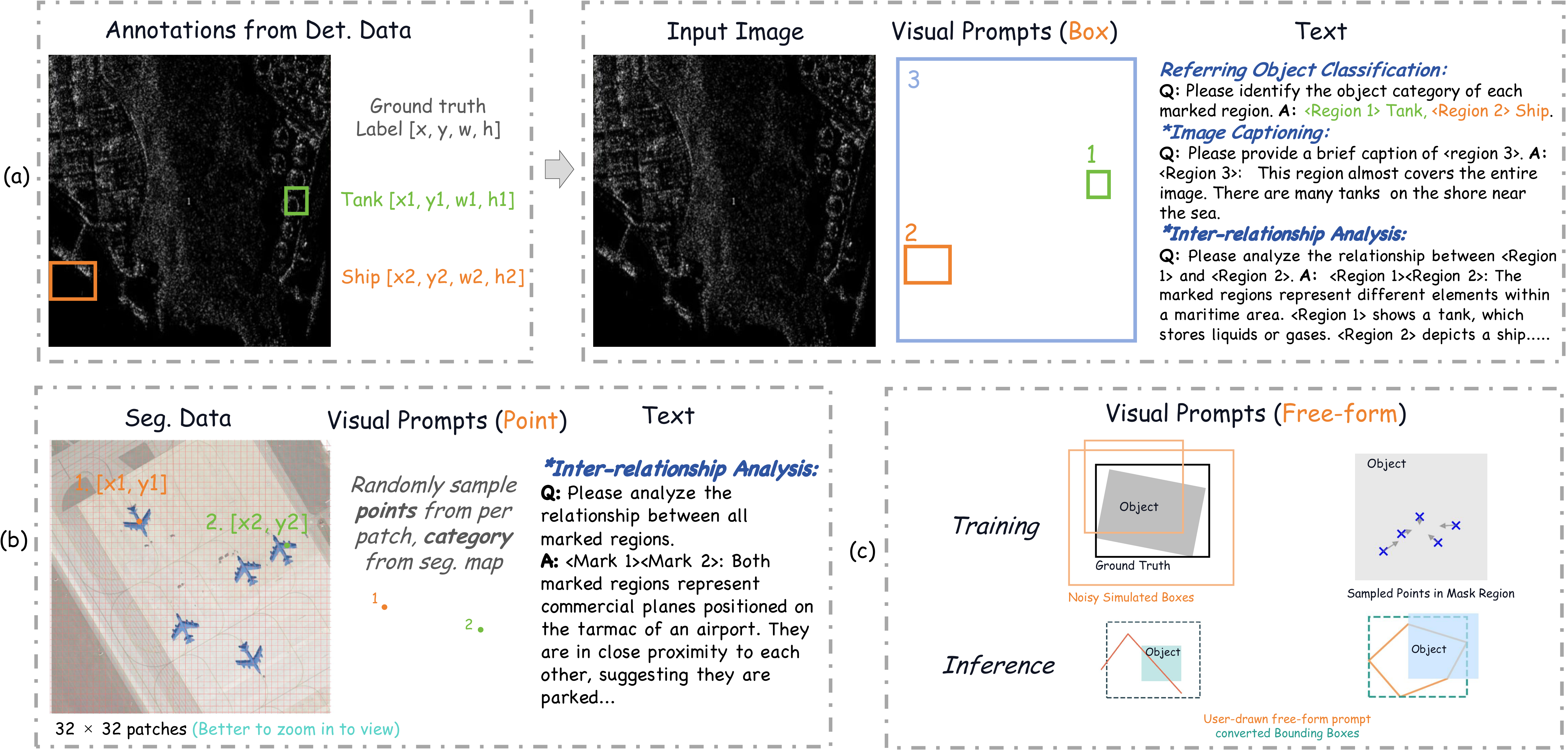}
	    \caption{(a) \textbf{Box prompts}: An example of converting SAR object detection data into multi-modal conversation training data. The training data consists of three parts: the image, the visual prompt, and the text (question and answer). The image is taken directly from the object detection dataset, while the visual prompts are from the ground-truth bounding boxes. The text prompts can be selected from fixed prompt templates (*note: enhanced by GPT-4V). (b) \textbf{Point prompts}: each image in the semantic segmentation dataset is divided into 32 × 32 patches, randomly sampled within each patch as the point prompts, with the category retrieved from the corresponding segmentation map. (c) \textbf{Free-form prompts}: Simulated by applying random noise-based augmentations to regions of interest. These free-form cues emulate flexible user interactions, enabling the model to generalize beyond structured visual prompts.
} 
	\label{FIG:Dataset}
\end{figure*}

\begin{figure*}[!h]
    \includegraphics[scale=0.175]{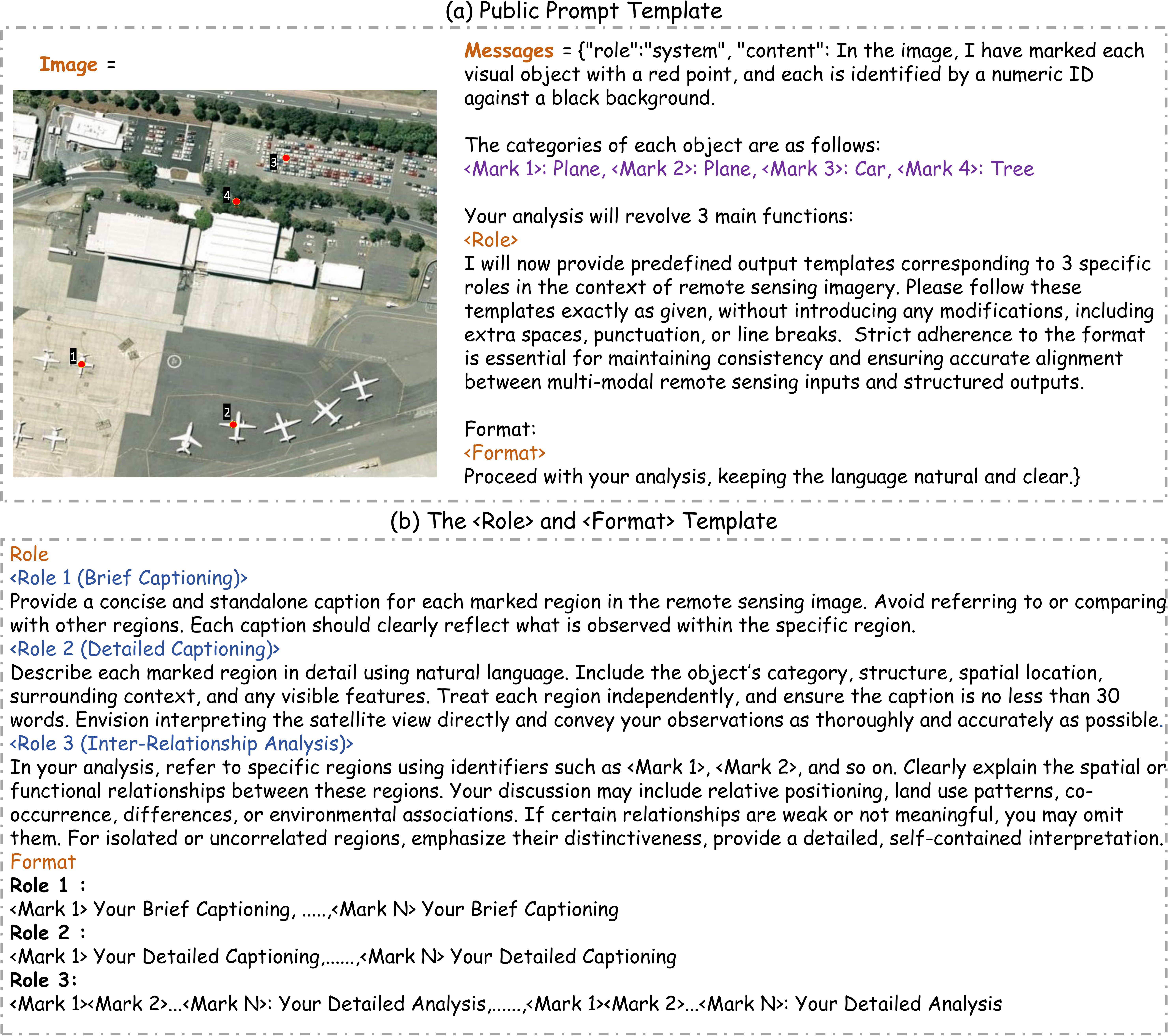}
	      \caption{Illustration of the standardized prompt templates designed for GPT-4V annotation generation. The (a) part shows the public templates for generic data construction, while the (b) part demonstrates the customized task templates, $<\mathrm{Role}>$ and $<\mathrm{Format}>$, tailored to multi-source RS imagery. These templates embed object categories explicitly, ensuring the correctness and consistency of generated annotations.}

	\label{FIG:Dataset2}
\end{figure*}

\section{Dataset Construction} 
To support comprehensive multi-source RS understanding, we construct a large-scale dataset called M-RSVP, which comprises over 650k samples spanning various tasks, modalities, and annotation formats. The detailed introduction is as follows.

\subsection{Multi-Source Multi-Domain Data Construction}
For developing multi-source and versatile image understanding abilities of EarthGPT-X, it is indispensable to utilize data from different scenes and sensors, including outdoor natural images, RS optical images, SAR, and infrared images, covering different visual task types. Thus, as shown in Tab.~\ref{tab:Dataset}, we construct a multi-domain collection of raw data covering diverse vision-language tasks. For the RS optical modality, we adopt RSVP~\cite{zhang2024earthmarker}, the first visual prompting dataset in the RS domain, which enables the model to learn region- and point-level understanding in complex geographic environments. Meanwhile, we incorporate more optical datasets~\cite{wang2023samrs,zhan2023rsvg} to enrich the diversity of tasks and enhance the model's generalization in RS scenarios. For SAR and infrared modalities, visual prompting data are mainly derived from publicly available object detection datasets, which are reorganized and annotated in a unified instruction format. Additionally, the natural scene dataset we selected is OpenImages~\cite{kuznetsova2020open} and Visual Genome~\cite{krishna2017visual}, which meets this characteristic and is ideal for training the model's visual perception capabilities in general environments. The multi-domain data integration ensures comprehensive and flexible image understanding across multiple domains and sensor types.
\subsection{Multi-level Prompting Data Annotation}
A meticulously designed multi-strategy annotation method is employed to label diverse multi-domain data, which is uniformly reorganized and formatted into \textit{image–visual prompt-text} triples. Notably, visual prompts include bounding boxes, points, and free-form types. The annotation process is detailed as follows.
\subsubsection{Box prompts}
The regional data are based on detection datasets, and ground truth bounding boxes are used as visual prompts to recognize object-level categories. As shown in Fig. \ref{FIG:Dataset} (a), an example of a SAR image is provided to illustrate the process. The traditional object detection data are transformed into multi-modal conversational samples, each comprising an image, visual prompts, and text (question-answer pairs). The image is directly sourced from the detection dataset, while the visual prompt is generated by applying slight perturbations to the ground-truth boxes, and the textual prompts are drawn from predefined templates. Specifically, image-level tasks use the boxes $[0, 0,\mathrm{width},\mathrm{height}]$, which can cover the entire image as the visual prompt. 
\subsubsection{Point prompts}
To enable point-based interaction during training, we construct point prompt annotations by randomly sampling pixels within each ground-truth object mask from semantic segmentation datasets. As illustrated in Fig. \ref{FIG:Dataset} (b), each image is divided into $32 \times 32$ patches, and the points are randomly sampled within each patch as the visual prompts,  where each point corresponds to a pixel-level label annotation.
\subsubsection{Free-form Prompts}
To enable the model to handle diverse and flexible visual inputs at inference time, we simulate free-form prompts through noise-based augmentation methods. Specifically, we consider both box and point prompts and augment them to reflect the imprecision and looseness of user-defined inputs, as illustrated in Fig.~\ref{FIG:Dataset} (c).
First, for box prompt augmentation, to mimic diverse user-drawn free-form inputs, we apply Gaussian noise to the original ground-truth bounding boxes. Let \( \mathcal{B} = (x, y, w, h) \) denote a bounding box parameterized by its top-left corner and dimensions. We generate an augmented box \( \mathcal{B}' \) via:
\begin{equation}
\mathcal{B}' = \mathcal{B} + \mathcal{N}(0, \Sigma), \quad \text{where } \Sigma \propto \text{diag}(w^2, h^2, w^2, h^2).
\end{equation}
This noise injection perturbs both the location and scale of the box in a size-aware manner, leading to boxes that may loosely cover, shift from, or extend beyond the target object, thus statistically approximating free-form visual prompts. 
For point prompts, we simulate user interactions by randomly sampling multiple pixels within the ground-truth object mask and associating them with the same instance label. Given a binary mask \( \mathcal{M} \subset \Omega \) representing a target object, we uniformly select \( K \) points from the interior of \( \mathcal{M} \). These sparse prompts guide the model to localize the target object via point-based reference.
Additionally, at inference (see bottom part of Fig.~\ref{FIG:Dataset} (c)), free-form inputs are pre-processed into bounding boxes, enabling flexible, intuitive user-drawn prompts.
\subsubsection{Text Prompts}
To support instruction tuning on multi-source RS data, we provide a set of carefully designed natural language instructions that correspond to different visual prompts (e.g., box or point). Each instruction guides the model to perform a specific visual understanding task, such as classification, captioning, or relational reasoning. Designed task instructions are as follows:
\begin{itemize}
    \item \textit{Please identify the object category of each marked region in the image.}
    \item \textit{Please identify the labels of each marked point in the image.}
    \item \textit{Please provide a brief caption of each marked region in the image.}
    \item \textit{Please provide a detailed caption of each marked point in the image.}
    \item \textit{Please analyze the relationship between all marked regions in the image.}
    \item \textit{Please provide a summarized caption based on all the marked regions in the image.}
    \item \textit{Please analyze how the marked objects interact with each other in the given scene.}
\end{itemize}
In addition, users can employ the $<$Mark n$>$ or $<$Region n$>$ identifiers to specify pixels or regions of interest and freely ask questions about them.
These task-specific textual instructions guide the model to perform multiple visual tasks, promoting a more fine-grained understanding of visual content under different granularities.
\subsection{GPT4V-assisted Data Generation}
To enable high-quality annotations with GPT-4V~\cite{openai2023gpt4}, we follows the design principles of~\cite{lin2024draw} and developed a set of rigorous and standardized prompt templates, as illustrated in Fig.~\ref{FIG:Dataset2}. These templates are tailored to distinct tasks: brief captioning, detailed captioning, and inter-relationship analysis. Notably, category names and ground-truth bounding boxes are explicitly embedded in the prompts, ensuring the correctness of the generated annotations by guiding GPT-4V to accurately associate text with visual regions. Beyond correctness, the structured templates also guarantee annotation consistency across large-scale data, avoiding ambiguity and hallucination.  
To further strengthen the coupling between visual prompts and textual outputs, we adopt Set-of-Marks (SoM) prompting~\cite{yang2023set}, which highlights multiple target objects simultaneously and reinforces spatial-textual grounding. This design enables GPT-4V to produce coherent and complete descriptions even in cluttered scenes. In conclusion, the integration of diverse multi-domain datasets through this rigorous pipeline allows joint training of EarthGPT-X on standardized supervision, enhancing its robustness across RS modalities and improving its generalization to complex real-world scenarios. Additionally, to further ensure reliability, we conduct RS expert inspection on sampled annotations, validating semantic accuracy, coherence, and completeness. This human-in-the-loop verification step complements automatic generation and safeguards the overall quality of the constructed dataset.

\section{Experiment}

\subsection{Experimental Setups}

\subsubsection{Implementation Details}
The one-stage all-in-one training is conducted in this paper. We use transformer-based decoder-only Llama 2-13B\cite{touvron2023llama_b} as the basic LLM. The pre-trained DINOv2-ViT L/14 \cite{oquab2023dinov2} and CLIP-ConvNeXt \cite{radford2021learning} are adopted as a hybrid image encoder MVE. The training image resolution is $448\times 448$. Subsequently, the randomly initialized ViT model is seated as the spatial-aware encoder to refine sparse features. Then, we employ a pre-trained SAM encoder and decoder\cite{kirillov2023segment} as the grounding encoder and pixel decoder. The image resolution required to be reshaped for the segmentation model is $1024\times1024$. Furthermore, we utilize the simple MLPs as the projection layer to realize multi-modal content dimension alignment. 

Throughout the training, to keep training efficiency and preserve the generalization ability of backbone models, we only unfreeze the self-attention layers of LLM and the added cross-attention module for tuning. Other modules, including visual encoders, projection layers, etc., are kept frozen. The AdamW optimizer\cite{kingma2014adam} with a weight decay of 0.01 and a learning rate of 2e-5 is utilized. The total training is conducted on 8 NVIDIA A100 GPUs with 80 GB memory, and the training time is around 310 hours.
\subsubsection{Evaluation Metrics}
We adopt three categories of evaluation metrics tailored to different visual tasks in our benchmark. 
\textit{(i) Region and Image Captioning Tasks:} We employ standard language generation metrics, including BLEU \cite{papineni2002bleu}, METEOR \cite{banerjee2005meteor}, ROUGE \cite{lin2004rouge}, CIDEr \cite{vedantam2015cider}, and SPICE \cite{anderson2016spice}. 
\textit{(ii) Scene Classification:} We report classification accuracy, which directly reflects the model’s ability to recognize high-level scene semantics in RS images.
\textit{(iii) Referring Object Classification:} We employ two semantic-aware metrics: Semantic Intersection over Union (S-IOU) and Semantic Similarity (SS) \cite{rezatofighi2019generalized,yuan2023osprey,lin2024draw}. SS measures the similarity of predicted and ground-truth labels in a semantic embedding space, while S-IOU captures the semantic overlap between them by computing the intersection-over-union in the embedding space.
These metrics provide a comprehensive assessment of the quality of the identified object information and generated captions. Consider the referring expression \textit{``An airplane is on the right of a small vehicle''}, 
where the ground-truth target labels are \textit{airplane} and \textit{vehicle}. 
Suppose the model predicts \textit{aircraft} and \textit{car}, both outputs achieve high SS since they are semantically close to the labels in the embedding space, and also high S-IoU because of strong token-level overlap. In contrast, if the predictions are \textit{ship} and \textit{building}, SS drops due to weaker semantic similarity, while S-IoU becomes very low as the key object words differ significantly from the ground-truth labels.

\subsection{Scene Classification}
We evaluate EarthGPT-X on two RS benchmarks: AID~\cite{xia2017aid} and UCMerced~\cite{yang2010bag}. Following GeoChat~\cite{kuckreja2024geochat}, we use 20\% of AID for testing, while the entire UCMerced serves as a zero-shot test set.
To fit the visual prompting setting, we define a full-image box $[0, 0, \mathrm{width}, \mathrm{height}]$ and use the instruction: ``Please identify the object category of each marked region in the image." Classification accuracy is computed under zero-shot settings. As shown in Tab.~\ref{tab:Classification_compare_zeroshot}, EarthGPT-X achieves 78.09\% on AID and 87.89\% on UCMerced, outperforming all baselines. This further confirms that the proposed visual prompting paradigm achieves stronger task alignment than directly applying vanilla MLLMs (e.g., Sphinx and GeoChat) without prompt adaptation.

\begin{table}[!t]
\caption{{Zero-shot Scene Classification (UCMerced and AID).}}
\label{tab:Classification_compare_zeroshot}
\centering
\renewcommand{\arraystretch}{1.2}
\scalebox{0.8}{
\fontsize{10pt}{10pt}\selectfont
\begin{tabular}{l c|cc}  
\toprule      
\multicolumn{1}{c}{Models} & \multicolumn{1}{c|}{{Publications}} & UCMerced Acc & AID Acc \\ 
\cmidrule(lr){1-4}
\multicolumn{1}{l}{Qwen-VL~\cite{bai2023qwen}} & {arXiv 2023} & 62.90  & 52.60  \\
\multicolumn{1}{l}{MiniGPTV2~\cite{chen2023minigpt}} & {arXiv 2023} & 4.76 & 12.90 \\
\multicolumn{1}{l}{LLaVa-1.5~\cite{liu2023visual}} & {NeurIPS 2023} & 68.00 & 51.00 \\
\multicolumn{1}{l}{Sphinx~\cite{lin2024sphinx}} & {ECCV 2024} & 62.76 & 58.20 \\
\multicolumn{1}{l}{InternVL~\cite{InternVL2_2024}} & {CVPR 2024} & 58.23 & 60.40 \\
\multicolumn{1}{l}{GeoChat~\cite{kuckreja2024geochat}} & {CVPR 2024} & 84.43 & 72.03 \\
\multicolumn{1}{l}{EarthMarker~\cite{zhang2024earthmarker}} & {TGRS 2025} & 86.52 & 77.97 \\
\cmidrule(lr){1-4}
\multicolumn{1}{l}{\textbf{EarthGPT-X (Ours)}} &  & \textbf{87.89} & \textbf{78.09} \\ 
\bottomrule
\end{tabular}
} 
\end{table}

\begin{table*}[!t]
\caption{Referring object classification (DIOR-RSVG).}
\label{tab:Referring}
\centering
\renewcommand{\arraystretch}{1.2}
\setlength{\tabcolsep}{6pt}
\scalebox{1.2}{
\begin{tabular}{l l|cc|cc|cc}
\toprule
\multirow{2}{*}{Models} & \multirow{2}{*}{{Publications}} & \multicolumn{2}{c|}{Point} & \multicolumn{2}{c|}{Box \textit{(Coor)}} & \multicolumn{2}{c}{Free-form} \\
\cmidrule(lr){3-4}\cmidrule(lr){5-6}\cmidrule(lr){7-8}
& & SS & S-IoU & SS & S-IoU & SS & S-IoU \\
\midrule
\multicolumn{8}{l}{\textit{\textbf{Vanilla MLLMs}}} \\
InternVL2.5~\cite{internlmxcomposer2_5_reward} & {arXiv 2024} & — & — & 54.54 & 37.91 & — & — \\
InternVL3~\cite{zhu2025internvl3}              & {arXiv 2025} & — & — & 50.84 & 26.99 & — & — \\
GeoChat~\cite{kuckreja2024geochat}             & {CVPR 2024}  & — & — & 79.59 & 68.80 & — & — \\
Sphinx~\cite{lin2024sphinx}                    & {ECCV 2024}  & — & — & 93.72 & 89.37 & — & — \\
EarthGPT~\cite{zhang2024earthgpt}              & {TGRS 2024}  & — & — & 94.64 & 90.16 & — & — \\
\midrule
\multicolumn{8}{l}{\textit{\textbf{Visual Prompting MLLMs}}} \\
Vip-LLava-7b~\cite{cai2024vip}                 & {CVPR 2024}  & 68.32 & 51.98 & 72.56 & 55.94 & 71.24 & 53.99 \\
Vip-LLava-13b~\cite{cai2024vip}                & {CVPR 2024}  & 70.02 & 57.61 & 74.51 & 60.53 & 74.05 & 59.83 \\
Sphinx\,-V~\cite{lin2024draw}                  & {ICLR 2025}  & 86.21 & 78.78 & 89.07 & 81.62 & 88.46 & 79.97 \\
EarthMarker~\cite{zhang2024earthmarker}        & {TGRS 2025}  & 95.96 & 93.49 & 98.37 & 97.24 & — & — \\
\midrule
\textbf{EarthGPT-X (Ours)}                      &           & {\textbf{96.33}} & {\textbf{94.42}} & \underline{{\textbf{98.79}}} & \underline{{\textbf{98.03}}} & {\textbf{98.71}} & {\textbf{96.72}} \\
\bottomrule
\end{tabular}}
\begin{center}
\scriptsize \textit{Notes: Since vanilla MLLMs lack visual prompt support, textual coordinates (abbr. as Coor) are treated as the referring regions. ‘—’ means no such capability. \\The \underline{underline} denotes the best performance achieved under box prompts.}
\end{center}
\end{table*}

\begin{table}[!b]
\caption{Referring Object Classification (MSAR).}
\label{tab:msar}
\centering
\renewcommand{\arraystretch}{1.2}
\scalebox{1.05}{
\begin{tabular}{l l |c|cc}
\toprule
Models & {Publications} & Formats & SS  & S-IoU  \\
\midrule
InternVL2.5~\cite{internlmxcomposer2_5_reward} & {arXiv 2024} & Coor & 37.05 & 7.69 \\
InternVL3~\cite{zhu2025internvl3}             & {arXiv 2025} & Coor & 32.72 & 2.89 \\
Sphinx-V~\cite{lin2024draw}                    & {ICLR 2025}  & Box  & 34.76 & 2.03 \\
\midrule
\textbf{EarthGPT-X (Ours)}                     &         & Box  & \textbf{99.85} & \textbf{97.75} \\
\bottomrule
\end{tabular}}
\end{table}

\begin{table*}[!hb]
\caption{Supervised Image Captioning (NWPU-captions). }
\label{tab:NWPU_caption_compare_supervised}
\centering
\renewcommand{\arraystretch}{1.2}
\setlength{\tabcolsep}{7pt}
\scalebox{0.8}{
\fontsize{11pt}{11pt}\selectfont
\begin{tabular}{l@{\hspace{1pt}}l |cccccccc}
\toprule
\multicolumn{1}{c}{Models} & \multicolumn{1}{c}{Publications} & BLEU-1 & BLEU-2 & BLEU-3 & BLEU-4 & METEOR & ROUGE\,-L & CIDEr(0-5) & SPICE \\ 
\cmidrule(lr){1-10}
\textit{\textbf{RS Expert Models}} & & & & & & & & & \\ 
CSMLF~\cite{8633358}                  & GRSL 2019 & 77.0 & 64.9 & 53.2 & 47.1 & 32.0 & 57.8 & 106.5 & 26.5 \\
Qu \textit{et al.}~\cite{qu2016deep} & CITS 2016 & 72.5 & 60.3 & 51.8 & 45.5 & 33.6 & 59.1 & 117.9 & 27.6 \\
Attention (soft)~\cite{lu2017exploring} & TGRS 2017 & 73.1 & 60.9 & 52.5 & 46.2 & 33.9 & 59.9 & 113.6 & 28.5 \\
Attention (hard)~\cite{lu2017exploring} & TGRS 2017 & 73.3 & 61.0 & 52.7 & 46.4 & 34.0 & 60.0 & 110.3 & 28.4 \\
FC-Att~\cite{zhang2019description}      & TGRS 2019 & 73.6 & 61.5 & 53.2 & 46.9 & 33.8 & 60.0 & 123.1 & 28.3 \\
SM-Att~\cite{zhang2019description}      & TGRS 2019 & 73.9 & 61.7 & 53.2 & 46.8 & 33.0 & 59.3 & 123.6 & 27.6 \\
MLCA-Net~\cite{cheng2022nwpu}          & TGRS 2022 & 74.5 & 62.4 & 54.1 & 47.8 & 33.7 & 60.1 & 126.4 & 28.5 \\ 
{Capformer~\cite{CapFormer}}             & {IGARSS 2022} & {89.0} & {81.1} & {74.5} & {69.2} & {45.4} & {78.4} & {201.9} & {--} \\
{Aware-Transformer~\cite{cao2023aware}} & {CGI 2023} & {\textbf{91.5}} & {\textbf{85.2}} & {\textbf{79.6}} & {\textbf{75.0}} & {\textbf{47.7}} & {\textbf{81.1}} & {\textbf{214.7}} & {--} \\
\cmidrule(lr){1-10}
\textit{\textbf{RS MLLMs}} & & & & & & & & & \\ 
{RS-CapRet~\cite{silva2024largelanguagemodelscaptioning}} & {arXiv 2024} & {87.1} & {78.7} & {71.7} & {65.6} & {43.6} & {77.6} & {192.9} & {31.1} \\
{EarthGPT~\cite{zhang2024earthgpt}}      & {TGRS 2024} & {87.1} & {78.7} & {71.6} & {65.5} & {44.5} & {78.2} & {192.6} & {\textbf{32.2}} \\
EarthMarker~\cite{zhang2024earthmarker} & TGRS 2025 & 84.4 & 73.1 & 62.9 & {54.3} & 37.5 & {70.0} & {162.9} & {26.8} \\
\midrule
\textbf{EarthGPT-X (Ours)}             &  & 85.2 & 74.2 & 64.0 & 55.9 & 39.5 & 71.5 & 163.7 & 27.3 \\
\bottomrule
\end{tabular}
}
\end{table*}

\begin{figure*}[!t]
		\includegraphics[scale=0.1678]{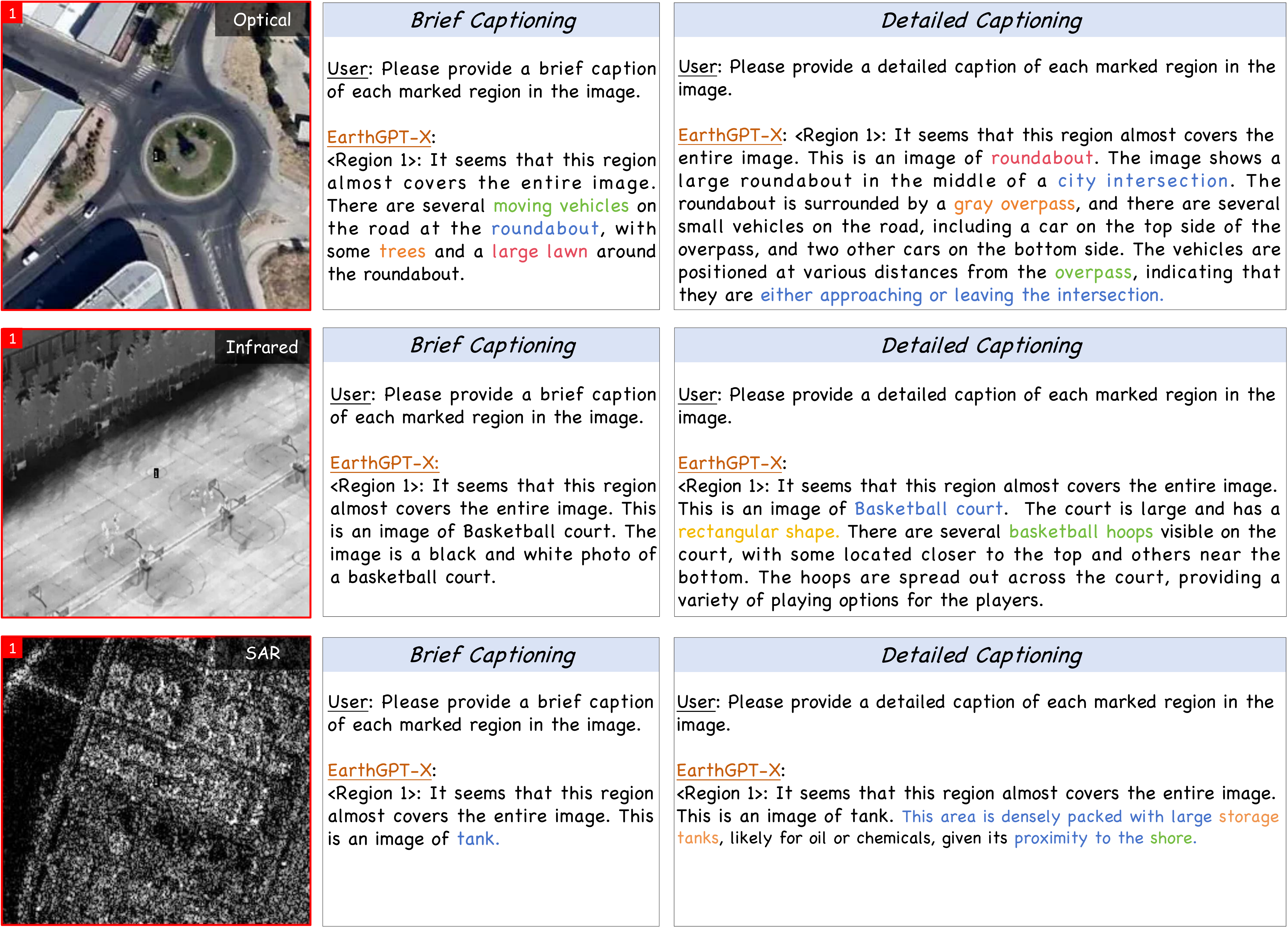}
	      \caption{Image brief and detailed captioning results on optical, infrared, and SAR modalities, demonstrating the superior scene-level visual understanding ability of EarthGPT-X.}
	\label{FIG:caption}
\end{figure*}

\begin{figure*}[!th]
	\centering
    \includegraphics[scale=0.12]{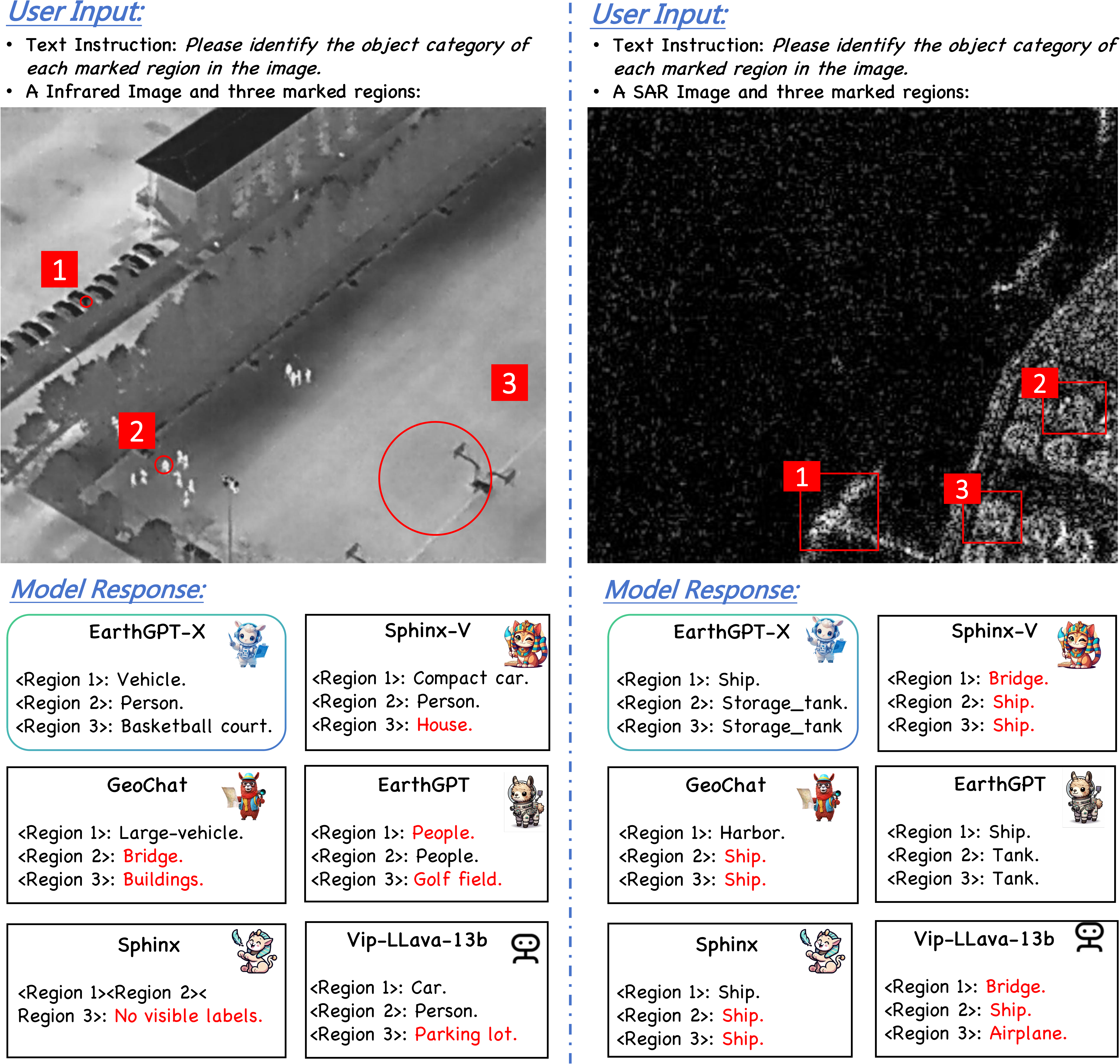}
	      \caption{The referring object classification results on SAR and infrared RS images demonstrate the superior object-level RS visual understanding capability of EarthGPT-X, compared to other vanilla and visual prompting MLLMs (\textit{The black text indicates consistency with the ground truth, the red text represents incorrect answers}).}
	\label{FIG:multi-source-compare}
\end{figure*}

\begin{table*}[!thb]
\caption{Zero-shot region captioning (OPT-RSVG).}
\renewcommand{\arraystretch}{1.2}
\centering
\scalebox{0.82}{
\fontsize{10pt}{10pt}\selectfont
\begin{tabular}{l@{\hspace{-6pt}} l l|cccccccc}  
\toprule      
\multicolumn{1}{c}{Models} & {Publications} & \multicolumn{1}{c}{{Format}} & BLEU-1 & BLEU-2 & BLEU-3 & BLEU-4 & METEOR & ROUGE & CIDEr & SPICE \\ 
\cmidrule(lr){1-3}\cmidrule(lr){4-11}
\textit{\textbf{Vanilla MLLMs}} & & & & & & & & & & \\       
Qwen-VL~\cite{bai2023qwen} & {arXiv 2023} & Coor & 12.78 & 6.47 & 2.77 & 1.24 & 5.97 & 17.06 & 13.15 & 6.60 \\
InternVL2.5~\cite{internlmxcomposer2_5_reward} & {arXiv 2024} & Coor & 9.87 & 4.18 & 1.69 & 0.52 & 5.71 & 10.55 & 9.22 & 8.28 \\
InternVL3~\cite{zhu2025internvl3} & {arXiv 2025} & Coor & 15.42 & 8.91 & 6.34 & 4.38 & 18.64 & 14.28 & 13.27 & 14.87 \\
GeoChat~\cite{kuckreja2024geochat} & {CVPR 2024} & Coor & 15.02 & 8.49 & 5.21 & 3.29 & 9.24 & 25.65 & 42.86 & 16.10 \\
Sphinx~\cite{lin2024sphinx} & {ECCV 2024} & Box & 22.79 & 15.27 & 10.97 & 7.84 & 12.95 & 30.28 & 105.85 & 20.18 \\
EarthGPT~\cite{zhang2024earthgpt} & {TGRS 2024} & Box & 30.28 & 19.99 & 13.58 & 8.73 & 15.29 & 31.52 & 88.92 & 21.05 \\ 
\cmidrule(lr){1-3}\cmidrule(lr){4-11}
\textit{\textbf{Visual Prompting MLLMs}} & & & & & & & & & & \\              
Sphinx-V~\cite{lin2024draw} & {ICLR 2025} & Box & 23.35 & 14.84 & 10.37 & 7.16 & 14.10 & 32.08 & 81.79 & 22.46 \\
Vip-LLava-7b~\cite{cai2024vip} & {CVPR 2024} & Box & 14.67 & 7.29 & 4.06 & 2.25 & 8.20 & 20.17 & 23.47 & 12.79 \\
Vip-LLava-13b~\cite{cai2024vip} & {CVPR 2024} & Box & 16.39 & 8.40 & 4.77 & 2.80 & 8.47 & 21.36 & 25.85 & 12.41 \\
GLaMM~\cite{rasheed2024glamm} & {CVPR 2024} & Box & 12.50 & 6.06 & 3.25 & 1.74 & 7.98 & 22.01 & 23.85 & 12.40 \\
\cmidrule(lr){1-3}\cmidrule(lr){4-11}
\textbf{EarthGPT-X (Ours)} &  & Point & {\textbf{52.57}} & {\textbf{48.14}} & {\textbf{45.72}} & {\textbf{43.38}} & {\textbf{35.28}} & {\textbf{64.63}} & {\textbf{479.92}} & {\textbf{63.47}} \\
\textbf{EarthGPT-X (Ours)} &  & Box & \underline{{\textbf{64.83}}} & \underline{{\textbf{59.91}}} & \underline{{\textbf{56.28}}} & \underline{{\textbf{53.64}}} & \underline{{\textbf{39.61}}} & \underline{{\textbf{66.87}}} & \underline{{\textbf{488.76}}} & \underline{{\textbf{64.32}}} \\
\textbf{EarthGPT-X (Ours)} &  & Free-form & {\textbf{63.38}} & {\textbf{58.62}} & {\textbf{55.12}} & {\textbf{52.41}} & {\textbf{38.72}} & {\textbf{65.43}} & {\textbf{483.63}} & {\textbf{63.58}} \\

\bottomrule
\end{tabular}
\label{tab:Region}
}
\begin{center}
\scriptsize \textit{Note: The \underline{underline} denotes the best zero-shot performance achieved under box prompts.}
\end{center}
\end{table*}

\begin{table*}[t]
\caption{Region Captioning on SAR and Infrared datasets.}
\centering
\renewcommand{\arraystretch}{1.2}
\scalebox{1}{
\fontsize{9pt}{9pt}\selectfont
\begin{tabular}{l l|ccc|ccc}  
\toprule
\multirow{2}{*}{Models} & \multirow{2}{*}{{Publications}} 
& \multicolumn{3}{c|}{SAR-Ship (Supervised)} 
& \multicolumn{3}{c}{HIT UAV (Zero-Shot)} \\
\cmidrule(lr){3-5}\cmidrule(lr){6-8}
&  & ROUGE-1 & ROUGE-L & METEOR & ROUGE-1 & ROUGE-L & METEOR \\
\midrule
\multicolumn{8}{l}{\textit{\textbf{Generic MLLMs}}} \\
GPT-4o~\cite{openai2024gpt4o}        & {OpenAI 2024} & 7.49  & 7.24  & 7.07  & 10.96 & 9.02 & 8.23 \\
InternVL2~\cite{InternVL2_2024}    & {OpenGV 2024}   & 9.67  & 8.67  & 8.19  & 11.00 & 9.53 & 8.40 \\
\midrule
\multicolumn{8}{l}{\textit{\textbf{Specialized RS MLLMs}}} \\
GeoChat~\cite{kuckreja2024geochat}   & {CVPR 2024}   & 57.15 & 57.15 & 52.20 & 59.85 & 59.85 & 51.31 \\
EarthDial~\cite{soni2025earthdial}   & {CVPR 2025}   & 63.10 & 63.10 & 54.83 & 61.83 & 61.83 & \textbf{52.80} \\
\midrule
\textbf{EarthGPT-X (Ours)}           &    & \textbf{65.25} & \textbf{65.25} & \textbf{55.90} & \textbf{64.40} & \textbf{64.40} & 51.56 \\
\bottomrule
\end{tabular}}
\label{tab:sar_caption}
\end{table*}

\begin{figure*}[!t]
	\centering
		\includegraphics[scale=0.17]{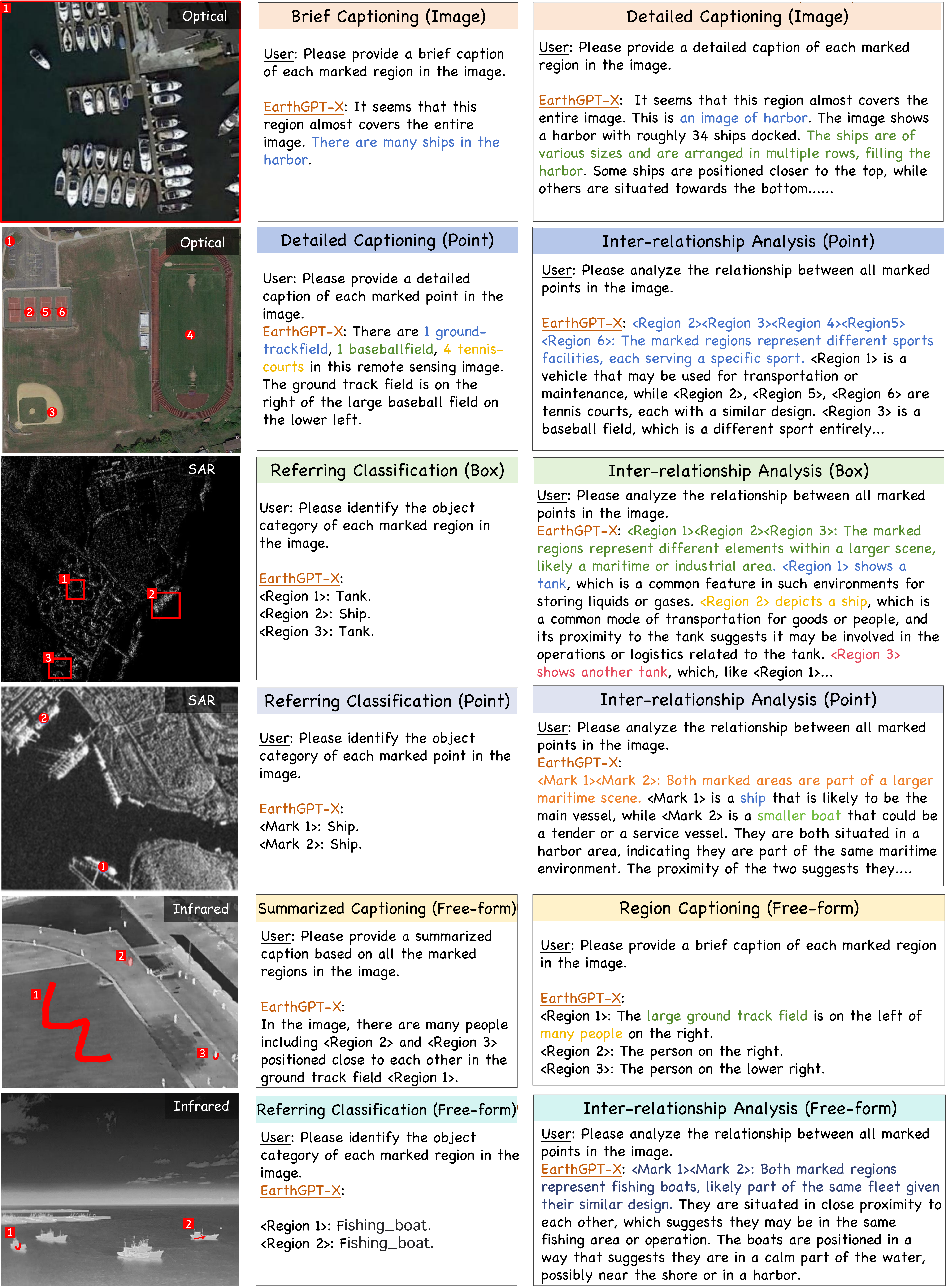}
	      \caption{Qualitative examples of EarthGPT-X performing multi-task reasoning and comprehending multi-source RS imagery (optical, SAR, and infrared) under various visual prompts.}
	\label{FIG:vis}
\end{figure*}

\begin{figure*}[!t]
	\centering
	\includegraphics[scale=0.2]{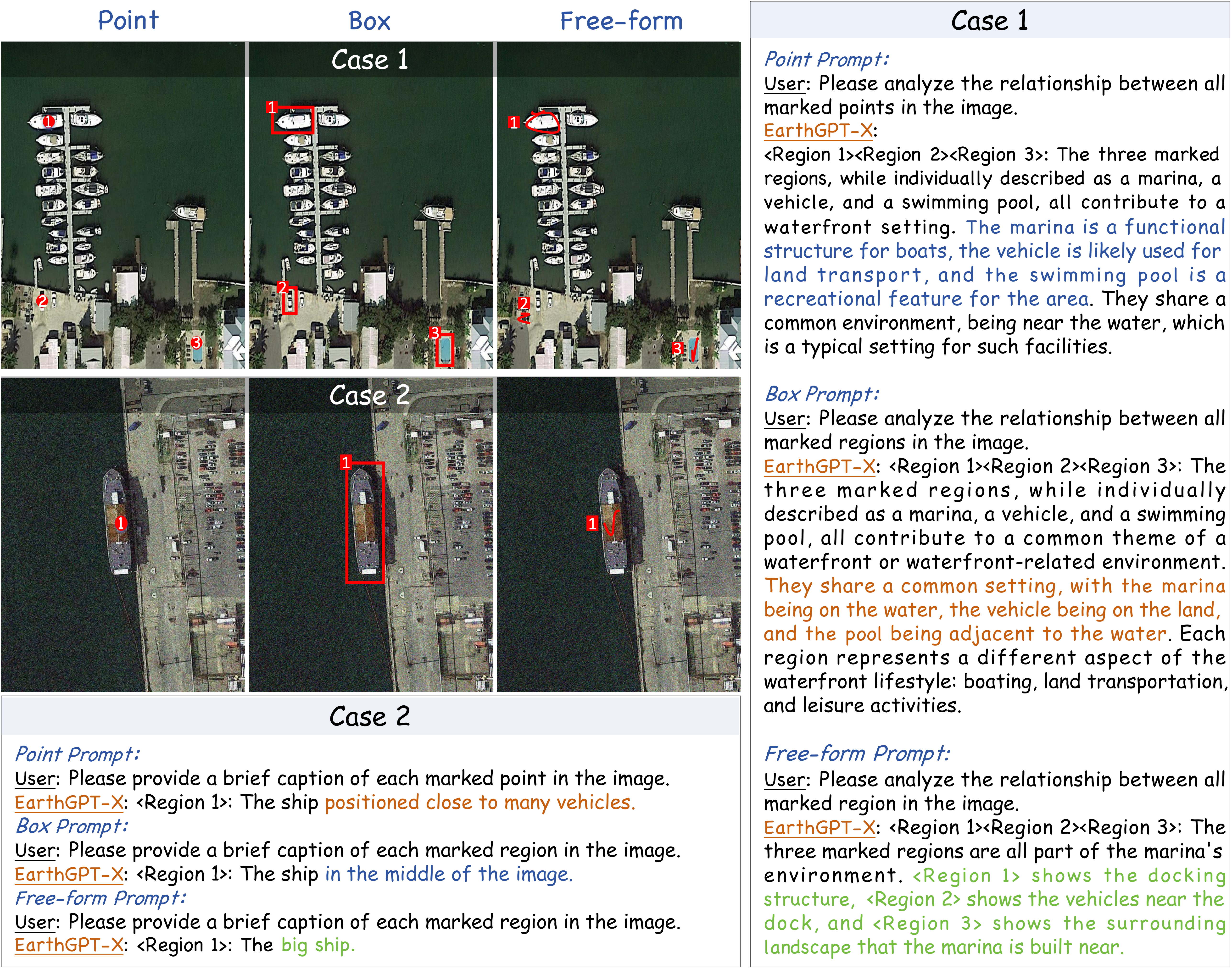}
	\caption{Illustration of multiple visual prompting strategies. Specifically, \textit{Point Prompt} highlights local functional roles of individual regions, \textit{Box Prompt} emphasizes global thematic coherence and spatial relations, and \textit{Free-form Prompt} provides concise contextual abstraction. Despite these differences, all three prompting types guide the model to establish precise region–text alignment. }
	\label{FIG:multi}
\end{figure*}

\begin{table*}[!h]
\centering 
\caption{{Ablation results about the design of the visual encoders. }}\label{tab:Ablation_on_visual_encoder}
\renewcommand{\arraystretch}{1.2}
\setlength{\tabcolsep}{2pt}
\scalebox{0.92}{
\fontsize{10pt}{10pt}\selectfont
\begin{tabular}{ccccccc}
\toprule  
\multicolumn{2}{c}{CNN Model}  & \multicolumn{1}{c|}{ViT Model}       & Classification             & Region Captioning            & Ref. Obj. Classification  \\ 
\cmidrule(lr){1-2}\cmidrule(lr){3-3}\cmidrule(lr){4-6}
CLIP ConvNeXt-L  &  CLIP RN50$\times$16 & \multicolumn{1}{c|}{DINOv2 ViT-L/14} & AID (Acc) & OPT-RSVG (CIDEr) & DIOR-RSVG (SIOU)  \\ 
\cmidrule(lr){1-2}\cmidrule(lr){3-3}\cmidrule(lr){4-6}
\checkmark             &   $\boldsymbol\times $            &      \multicolumn{1}{c|}{$\boldsymbol\times $ }             & 73.65                     & 471.02           & 94.11         \\
$\boldsymbol\times $   & \checkmark                        &      \multicolumn{1}{c|}{$\boldsymbol\times $ }              & 74.26                    & 471.35           & 94.68         \\
$\boldsymbol\times $   & $\boldsymbol\times $              & \multicolumn{1}{c|}{\checkmark}              & 73.05               & 479.12                 & 96.07         \\
 $\boldsymbol\times $  & \checkmark                        & \multicolumn{1}{c|}{\checkmark}              & 75.51               & 480.63                 & 96.92        \\
\checkmark             &  $\boldsymbol\times $             & \multicolumn{1}{c|}{\checkmark}              & \textbf{78.09}                     & \textbf{488.76}           & \textbf{98.03}         \\ 
\bottomrule
\end{tabular}
}\label{ab}
\begin{center}
\scriptsize \textit{Notes: Results for region-level tasks are reported under box-prompt settings.}
\end{center}
\end{table*}

\subsection{Referring Object Classification}
\subsubsection{Evaluation on Optical Data}
Referring object classification aims to identify the category labels of objects referred to by visual prompts and textual instructions. To assess the merits of the proposed EarthGPT-X, we report its performance and compare it with the recent vanilla and visual prompting MLLMs on DIOR-RSVG\cite{zhan2023rsvg}. We adapt DIOR-RSVG by converting referring expressions and their bounding boxes into classification samples, where each target region serves as input and its category label as output.
The text instruction is ``Please identify the object category of the marked region in the image". The vanilla MLLMs do not support drawing visual prompts on the images, therefore, the coordinates contained in the text instruction serve as the referring regions. The vanilla MLLMs used for comparison include natural-domain models such as Sphinx~\cite{lin2024sphinx}, InternVL2.5\cite{internlmxcomposer2_5_reward}, InternVL3\cite{zhu2025internvl3}, as well as RS-domain models including GeoChat~\cite{bai2023qwen} and EarthGPT~\cite{zhang2024earthgpt}.
For visual prompting MLLMs, we select the most recent representatives, namely the natural-domain models Sphinx-V~\cite{lin2024draw} and Vip-LLava~\cite{cai2024vip}, together with the RS-domain model EarthMarker~\cite{zhang2024earthmarker}.
We conduct experiments under three levels of visual prompting (point, box, and free-form) to provide a fair and comprehensive comparison across models.
Following~\cite{lin2024draw, you2023ferret}, for point prompts, we randomly select a point within the ground-truth object that is also close to its boundary. For box prompts, we use the ground-truth bounding box provided by DIOR-RSVG. For free-form prompts, we randomly perturb the center positions and zoom the box prompts to simulate.
As shown in Tab. \ref{tab:Referring}, EarthGPT-X achieves optimal results in S-IoU under box prompts. We also observe that the box-level results are better than the point and free-form prompt ones.

\subsubsection{Evaluation on SAR Data}
We evaluate EarthGPT-X and recent InternVL series\cite{internlmxcomposer2_5_reward,zhu2025internvl3}, Sphinx-V\cite{lin2024draw} on the MSAR\cite{chenlarge} test set (Tab.~\ref{tab:msar}). EarthGPT-X, trained with SAR-specific samples, achieves superior performance (SS: 99.85\%, S-IoU: 97.75\%), showing strong capacity in SAR-oriented understanding. In contrast, natural MLLMs InternVL2.5 and InternVL3, without SAR training, perform significantly lower. The superior results of EarthGPT-X illustrate its stronger modality adaptation and specialization capabilities, which are crucial for accurate SAR imagery understanding in the RS applications.

\subsection{Image Captioning}
To evaluate the image captioning capabilities, we adopt the NWPU-Captions~\cite{cheng2022nwpu} dataset to benchmark EarthGPT-X against other expert models under supervised settings. During evaluation, we use the instruction ``Please provide a brief caption of each marked region in the image" alongside a full-image box $[0, 0,\mathrm{width},\mathrm{height}]$ as the visual prompt to guide the captioning process. \blue{ As shown in Table~\ref{tab:NWPU_caption_compare_supervised}, EarthGPT-X outperforms most expert-designed models but remains inferior to task-specific architectures such as Aware-Transformer~\cite{cao2023aware}. We note that the performance gap in image captioning is a common limitation of MLLMs, which aim to balance multiple capabilities rather than optimizing for a single task.}

\subsection{Region Captioning}

\subsubsection{Evaluation on Optical Data} For the brief region captioning on optical data, the test set of OPT-RSVG~\cite{li2024language} is employed for zero-shot evaluation. We elaborately choose both natural and RS-specific MLLMs for performance comparison. Natural scene MLLMs contain  Qwen-VL~\cite{kuckreja2024geochat}, InternVL2.5\cite{internlmxcomposer2_5_reward}, InternVL3\cite{zhu2025internvl3}, Sphinx~\cite{lin2024sphinx}, Sphinx-V~\cite{lin2024draw}, Vip-LLava~\cite{cai2024vip} and GLaMM~\cite{cai2024vip}. Popular RS-specific MLLMs GeoChat~\cite{bai2023qwen} and EarthGPT~\cite{zhang2024earthgpt} are selected. The text instruction for the region captioning task is designed as ``Please provide a brief caption of each marked region in the image". As shown in Tab.~\ref{tab:Region}, the proposed EarthGPT-X consistently outperforms previous works across all evaluation metrics. Notably, EarthGPT-X achieves significant gains over Sphinx-V, despite the latter being trained on large-scale RS datasets. This highlights the superior performance and stronger generalization ability of EarthGPT-X in multi-source RS data understanding. Notably, using boxes as visual prompts can yield the best performance on region captioning ability.

\subsubsection{Evaluation on SAR and Infrared Data}

As displayed in Tab.~\ref{tab:sar_caption}, for the multi-source brief captioning task, 
we evaluate EarthGPT-X on two representative datasets: 
SAR-Ship~\cite{rs11070765} for supervised evaluation 
and HIT UAV~\cite{suo2023hit} under a zero-shot setting. 
These datasets enable us to validate the generalization ability of EarthGPT-X across different modalities. 
For comparison, we include two representative general-purpose MLLMs, 
GPT-4o from OpenAI and InternVL2~\cite{InternVL2_2024} from OpenGVLab. 
We further reference the first RS MLLM, GeoChat~\cite{kuckreja2024geochat}, 
and the latest multi-source understanding model, EarthDial~\cite{soni2025earthdial}, 
as strong domain-related baselines. 
Experimental results demonstrate that EarthGPT-X achieves superior overall performance on multi-source datasets.

\subsection{Ablation Study}

\begin{table}[!b]
\centering 
\caption{Ablation results on Image Pre-processing.}\label{tab:Ablation on Image Pre-processing}
\renewcommand{\arraystretch}{1.5}
\setlength{\tabcolsep}{7pt} 
\scalebox{1}{
\fontsize{8.6pt}{8.6pt}\selectfont
\begin{tabular}{cccc}
\toprule  
\multicolumn{1}{c}{Settings} & \multicolumn{1}{c|}{Visual Encoder} & SS &  S-IoU \\ 
\cmidrule(lr){1-2}\cmidrule(lr){3-4}
\multicolumn{1}{c}{Single-scale}  & \multicolumn{1}{c|}{MVE} & 94.93 & 95.05         \\
\multicolumn{1}{c}{Multi-scale}   & \multicolumn{1}{c|}{MVE} & 96.18 & 95.29          \\
\multicolumn{1}{c}{Multi-scale + Sub-images} & \multicolumn{1}{c|}{MVE} & \textbf{98.79} & \textbf{98.03}         \\ 
\bottomrule
\end{tabular}
}
\begin{center}
\scriptsize \textit{Notes: Results are reported under box-prompt settings.}
\end{center}
\end{table}

\begin{table}[!h]
\caption{Computation Analysis and Comparisons}
\label{tab:Inference Costs}
\renewcommand{\arraystretch}{1.1}
\scalebox{0.84}{
\fontsize{10.3pt}{10.3pt}\selectfont
\begin{tabular}{lcc}
\toprule
\multicolumn{1}{c}{\!\!\!\! Models\!\!\!\!\!\!\!\!\!\!\!\!}&  \!\! Inference Time(s) &\!\!\!\!\! GPU Memory(MiB) \\ 
\cmidrule(lr){1-1}\cmidrule(lr){2-3}
\textit{\textbf{MLLMs}}          \\ 
\multicolumn{1}{l}{Sphinx}        & 1.08 & 41853              \\
\multicolumn{1}{l}{EarthGPT}      & 0.79 & 41766              \\
\multicolumn{1}{l}{GeoChat}       & 0.74 & 15544              \\
\multicolumn{1}{l}{Qwen-VL}  & 0.35 & 38145              \\
\cmidrule(lr){1-1}\cmidrule(lr){2-3}
\textit{\textbf{Visual Prompting Models}}          \\ 
\multicolumn{1}{l}{Sphinx-V}            & 1.95 & 41794      \\
\multicolumn{1}{l}{Vip-LLava-13B}       & 0.84 & 27748      \\
\multicolumn{1}{l}{{EarthGPT-X(Ours)}} & {1.60} & {41006} \\ 
\bottomrule
\end{tabular}
}
\end{table}

\subsubsection{Different Visual Encoders} To show the effectiveness of the proposed hybrid MVE module, we ablate on the performance of our EarthGPT-X when using different visual encoders. Specifically, this ablation uses CLIP-ConvNeXt-L, CLIP RN50×16, and DINOv2 ViT-L/14 for comparison. The results are shown in Tab. \ref{ab}, it is clear that the setting using the CLIP-ConvNeXt-L and DINOv2 ViT-L/14 as the mixed visual experts achieves the most superior performance on scene classification, region captioning, and referring object classification tasks. These results demonstrate that utilizing a network architecture with complementary functions has deepened the understanding of local information and contextual dependencies on RS imagery.

\subsubsection{Sub-images Decomposition and Multi-scale} 
In the visual feature extraction stage, we strengthen feature integration by leveraging multi-scale features and high-resolution sub-images for richer RS image comprehension. Ablation results in Tab.~\ref{tab:Ablation on Image Pre-processing} show that this design significantly outperforms single-scale settings under the referring object classification task, confirming that capturing finer details through multiple scales and sub-images is crucial for accurate interpretation.

\subsection{Computation Analysis and Comparisons}
In this section, we compare the GPU memory usage (MiB) and inference time (seconds) of EarthGPT-X with other MLLMs and visual prompting models. All evaluations are conducted on the region captioning task using an NVIDIA RTX A6000. For both MLLMs and visual prompting models, GPU memory usage is primarily attributed to model weight storage. As illustrated in Tab.~\ref{tab:Inference Costs}, EarthGPT-X consumes more GPU memory than Sphinx~\cite{lin2024draw}, GeoChat~\cite{kuckreja2024geochat}, EarthGPT~\cite{zhang2024earthgpt}, Qwen-VL~\cite{bai2023qwen}, and Vip-LLava-13B~\cite{cai2024vip}, but less than Sphinx-V~\cite{lin2024draw}. Similarly, its inference time is longer than all other models except Sphinx-V, due to its larger parameter scale. However, the time difference remains within one second, indicating that the performance gap is marginal and does not affect deployment in practical applications, as EarthGPT-X remains well within the capacity of commonly used GPUs. In future work, we plan to optimize EarthGPT-X’s inference efficiency to better balance speed and interpretability.

\begin{figure*}[!t]
	\centering
		\includegraphics[scale=0.136]{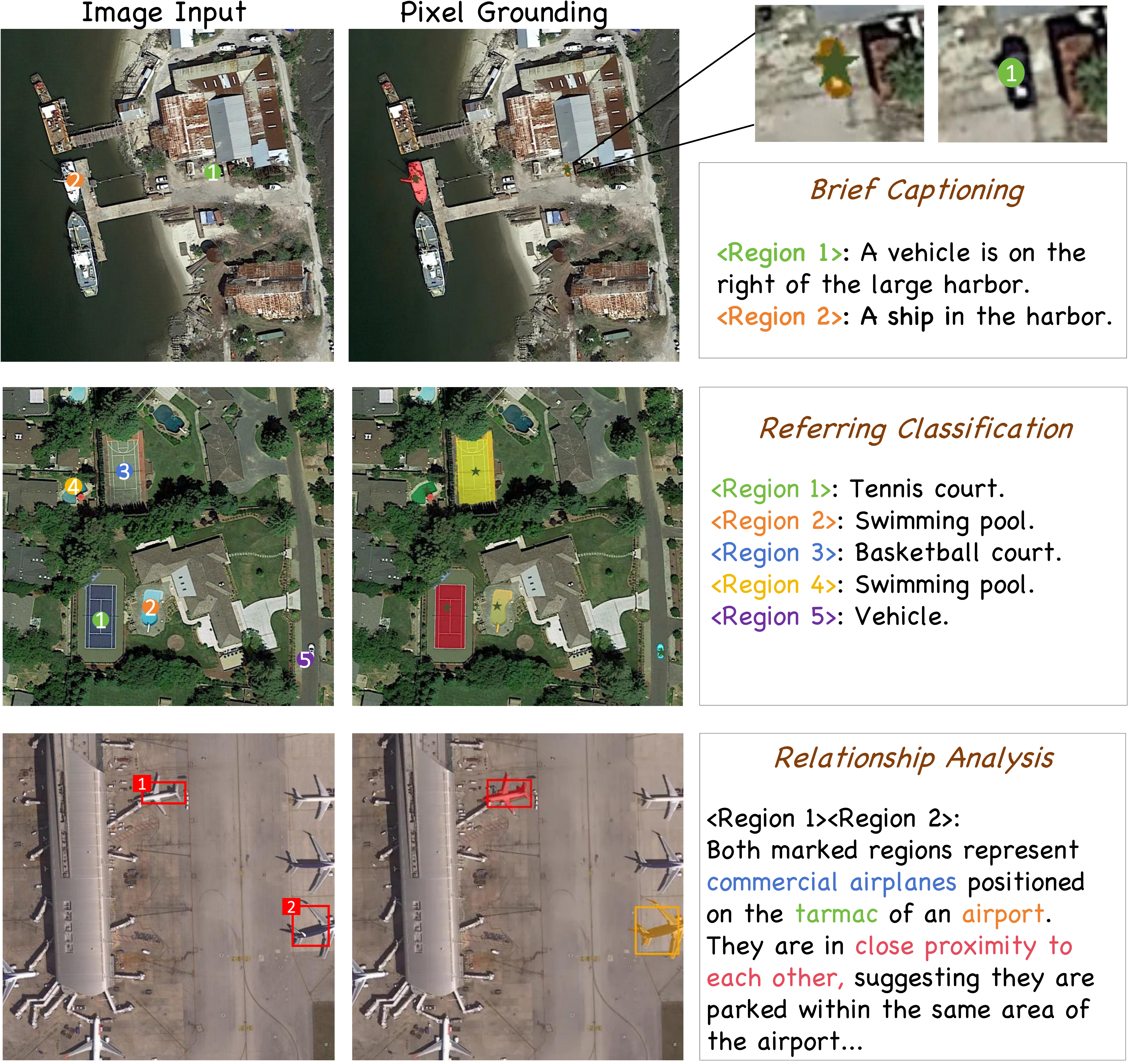}
	      \caption{Examples of EarthGPT-X performing visual reasoning and pixel-level grounding on optical imagery across diverse scenes (harbor, urban stadium, and airport). \textit{\textbf{Better to zoom in to view clearly}}.}
	\label{FIG:pixel}
\end{figure*}

\begin{figure*}[!t]
	\centering
	\includegraphics[scale=0.19]{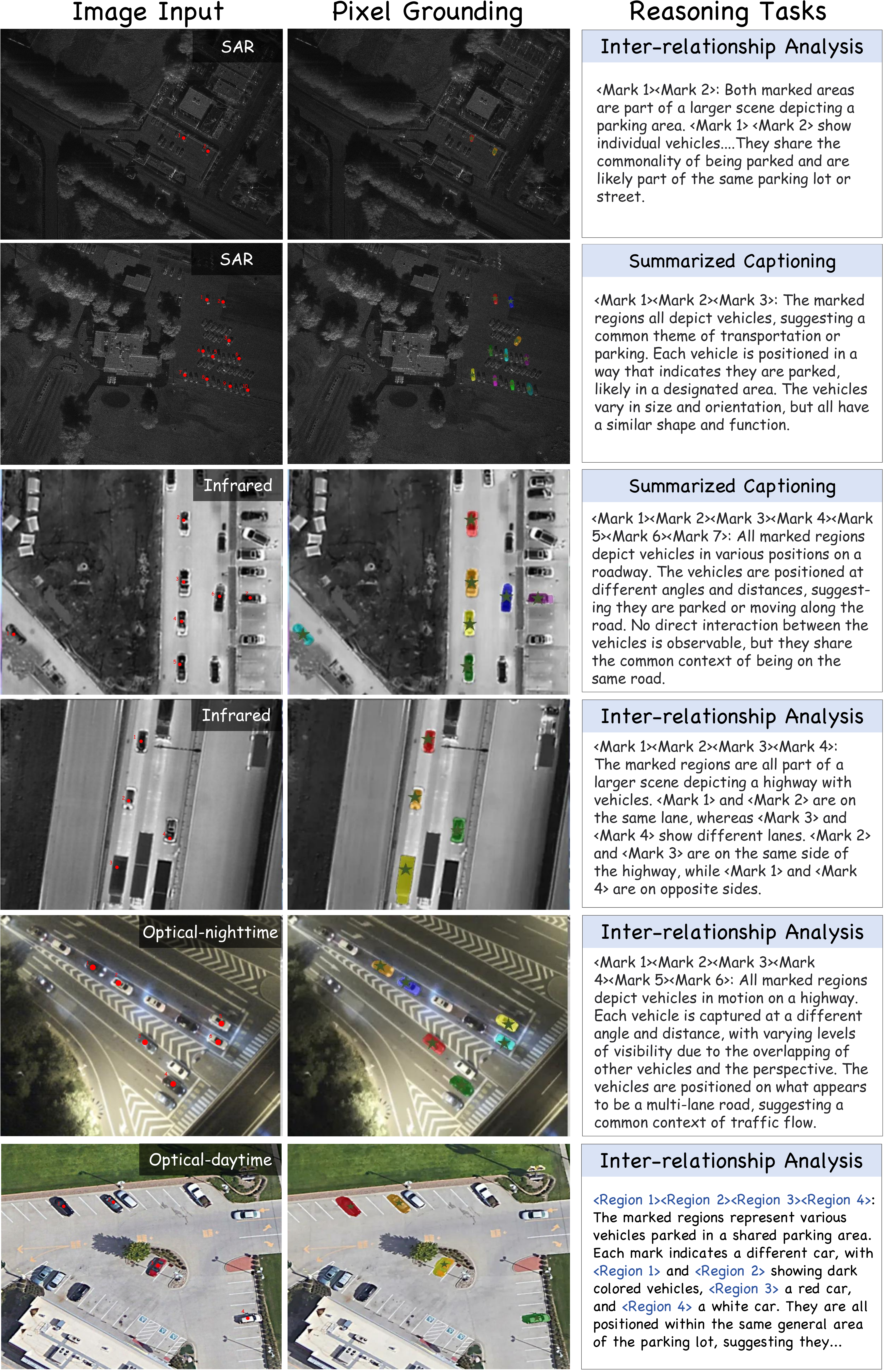}
	      \caption{Examples of EarthGPT-X performing small vehicle recognition and reasoning with pixel-level grounding on multi-source imagery (i.e., SAR, infrared, optical–daytime, and optical–nighttime). \textit{\textbf{Better to zoom in to view clearly.}}}
	\label{FIG:pixel2}
\end{figure*}

\section{Visualization}

\subsection{Multi-source Image Captioning}
We present visualization results of EarthGPT-X on optical, SAR, and infrared images to demonstrate its capability in understanding diverse RS scenes. Given an entire image as the region of interest, we use the visual prompt $[0, 0, \mathrm{width}, \mathrm{height}]$ and provide instructions such as ``Please provide a brief caption of each marked region in the image.
'' or ``Please provide a detailed caption of each marked region in the image.'' As shown in Fig.~\ref{FIG:caption}, the model generates accurate and semantically rich captions across modalities, effectively handling varied visual characteristics and noise patterns. 

\subsection{Referring Object Classification}
We conduct qualitative analysis to provide an intuitive performance display of EarthGPT-X. As shown in Fig. \ref{FIG:multi-source-compare}, the infrared image showcases a residential leisure area, and for the three marked regions, all other MLLMs respond incorrectly. Conversely, EarthGPT-X successfully recognizes the human, vehicle, and basketball court targets. Following this is a fuzzy port SAR image, 
where it is difficult to visually distinguish between ships and storage tanks. EarthGPT-X delivers precise interpretations for these targets, whereas other models perform poorly, except for EarthGPT. These results prove the significance of our work that empowers the MLLMs with fine-grained reasoning ability for object-level comprehension.

\subsection{Multi-level Multi-task Reasoning}
Traditional vision models in the RS field primarily focus on a single visual modality and single tasks, exhibiting poor generalization\cite{zhang2023posterior,zhuang2024heterogeneous,10443257}. Thanks to our M-RSVP's rich multi-sensor knowledge, EarthGPT-X has robust generalization capabilities across optical, SAR, and infrared visual modalities. To verify and assess the comprehensive capabilities of the proposed model, we conduct a thorough qualitative evaluation. Our assessment also covers multi-source images and multiple levels of image interpretation granularity, including image, point, box, and free-form prompted levels. Moreover, the evaluated visual tasks are extensive, such as brief captioning, detailed captioning, ground captioning, region captioning, referring object classification, and inter-relationship analyses. As displayed in Fig. \ref{FIG:vis}, our model demonstrates remarkable visual reasoning and analytical abilities. EarthGPT-X provides detailed descriptions for multi-source images, as evidenced by the interpretation of the first image, which includes overall scene depiction, target counts, orientations, and higher-level reasoning. In the relationship analysis of the second and third examples, the model generates comprehensive and structured responses, effectively capturing both local and global semantics. Notably, even for low-quality SAR and infrared maritime images, EarthGPT-X can infer relative spatial relations, target sizes, vessel functionalities, and operational states (e.g., docking at a port). In summary, EarthGPT-X demonstrates a profound ability to dissect and understand complex visual information in the multi-source RS MLLMs.

\subsection{Multiple Visual Prompts}
To evaluate the adaptability of EarthGPT-X under different prompting shapes, we investigate point prompt, box prompt, and free-form prompt. These prompts vary in the way visual regions are highlighted to the model, thereby influencing the style and granularity of responses. \textbf{\textit{i) Similarities}:} Across all prompt types, EarthGPT-X consistently grounds the responses to the marked regions, ensuring semantic coherence and correct alignment with the visual input.  \textbf{\textit{ii) Differences}}: As shown in Fig. \ref{FIG:multi}, under the point prompt, EarthGPT-X generates a caption emphasizing the spatial relation, which reflects its local context-awareness. Under the box prompt, the model highlights the absolute spatial location, showing a region-level grounding behavior. In contrast, the free-form prompt yields a more generic description, focusing on object-level attributes (\textit{Note that these examples are qualitative illustrations, not unique or deterministic results, since different prompt placements may lead to varied outputs}).

\subsection{Zero-shot Pixel-level Grounding and Reasoning}
All pixel grounding and reasoning experiments are conducted in a zero-shot setting, and the capability is obtained without any task-specific training.
\subsubsection{Different Scenes}
To evaluate the robustness under complex scenes, we select RS images with multiple diverse targets. As shown in Fig.~\ref{FIG:pixel}, in the port scene, we mark a ship and a tiny vehicle to examine the model's ability to localize small objects within a cluttered background. EarthGPT-X not only successfully detects these targets but also separates them from the surroundings. In another urban sports center scene, five spatially separated objects (e.g., swimming pool, tennis court, and vehicle) are interpreted. EarthGPT-X demonstrates exceptional performance in precise object recognition and pixel-grounding.
\subsubsection{Different Image Modalities}
As shown in Fig.~\ref{FIG:pixel2}, we further evaluate on small vehicle targets across heterogeneous image modalities, including optical, infrared, and SAR. The test data is from VisDrone-DroneVehicle~\cite{sun2020drone} and SAR\_Vehicle\_Detection\_Dataset\cite{SARV}. Despite the significant domain gaps among these modalities, EarthGPT-X consistently identifies and segments small and tiny vehicles, showing its strong generalization and adaptability to heterogeneous RS image types.

\section{Conclusion}

In this paper, we propose EarthGPT-X, a flexible and versatile MLLM specifically designed for understanding RS imagery from multiple sources. Technically, the hybrid signal mutual understanding mechanism is developed to enhance the interplay between dense visual features, sparse free-form visual prompts, and text instructions, facilitating a multi-grained understanding of complex RS imagery. Furthermore, a novel cross-domain one-stage fusion training strategy is proposed to generalize general-domain knowledge into RS multi-source scenarios. Through the designed visual prompt learning framework and training strategy, EarthGPT-X achieves multi-level visual understanding at the image, region, and pixel levels, etc., allowing for precise and intelligent analyses in real-world applications. \blue{In the future, we plan to enhance the image captioning capability of EarthGPT-X through task-aware fine-tuning and adaptive prompt optimization.} Furthermore, 
we aim to expand EarthGPT-X towards a wider range of tasks and more advanced visual reasoning capabilities across diverse modalities, ultimately strengthening its versatility and effectiveness in comprehensive RS imagery analysis. 

\bibliographystyle{unsrt}
\bibliography{my.bib}

\end{document}